\documentclass[twocolumn, 10pt]{article}
\usepackage[letterpaper, textheight=9.25in, textwidth=43pc, top=0.75in, columnsep=1pc, headheight=12pt, headsep=18pt]{geometry}
\usepackage{float}
\usepackage{multirow}
\usepackage{nicematrix}
\usepackage{tikz}
\usepackage{amsmath}
\usepackage{amsfonts}
\usepackage{xcolor}
\usepackage{booktabs}
\usepackage{adjustbox}
\usepackage{array}
\newcolumntype{H}{>{\setbox0=\hbox\bgroup}c<{\egroup}@{}}
\usepackage{algorithmic}
\usepackage{algorithm}

\usepackage{array}
\usepackage[caption=false,font=footnotesize]{subfig}
\usepackage{url}
\usepackage{hyperref}
\usepackage[capitalize,noabbrev]{cleveref}
\usepackage{enumitem}
  \newlist{inlinelist}{enumerate*}{1}
  \setlist*[inlinelist,1]{%
          label=(\roman*),
      }
\newtheorem{theorem}{Theorem}[section]

\newcommand{\data}{\mathcal{D}}

\newcommand{\numbershadowmodels}{M}

\newcommand{\hpofunction}{$\texttt{HPO}$}
\newcommand{\trainfunction}{$\textsc{train}$}

\newcommand{\model}{\mathcal{M}} 
\newcommand{\modeltar}{\mathcal{M}_{\mathcal{T}}} 
\newcommand{\modelshadow}{\mathcal{M}_{\mathcal{S}}}

\newcommand{\tpr}{\textsc{tpr}}
\newcommand{\fpr}{\textsc{fpr}}

\newcommand{\architecture}{\mathcal{A}}
\newcommand{\prob}{\mathbb{P}}
\newcommand{\normal}{\mathcal{N}}
\newcommand{\attack}{\texttt{KNOWN}}
\newcommand{\bb}{\texttt{BLACK-BOX}}
\newcommand{\logits}{\textsc{logits}}

\begin{document}

\title{Hyperparameters in Score-Based Membership Inference Attacks
\thanks{This work has been accepted for publication in the 3rd IEEE Conference on Secure and Trustworthy Machine Learning (SaTML'25). The final version will be available on IEEE Xplore.}}
\date{}
\author{
  \textbf{Gauri Pradhan}\\University of Helsinki, Finland \\ \texttt{gauri.pradhan@helsinki.fi}
  \and \textbf{Joonas J\"{a}lk\"{o}}\\University of Helsinki, Finland \\ \texttt{joonas.jalko@helsinki.fi}
  \and \textbf{Marlon Tobaben}\\University of Helsinki, Finland \\ \texttt{marlon.tobaben@helsinki.fi}
  \and \textbf{Antti Honkela}\\University of Helsinki, Finland \\ \texttt{antti.honkela@helsinki.fi}
}

\maketitle

\begin{abstract}
Membership Inference Attacks (MIAs) have emerged as a valuable framework for evaluating privacy leakage by machine learning models. Score-based MIAs are distinguished, in particular, by their ability to exploit the confidence scores that the model generates for particular inputs. Existing score-based MIAs implicitly assume that the adversary has access to the target model's hyperparameters, which can be used to train the shadow models for the attack. 
In this work, we demonstrate that the knowledge of target hyperparameters is not a prerequisite for MIA in the transfer learning setting. 
Based on this, we propose a novel approach to select the hyperparameters for training the shadow models for MIA when the attacker has no prior knowledge about them by matching the output distributions of target and shadow models. We demonstrate that using the new approach yields hyperparameters that lead to an attack near indistinguishable in performance from an attack that uses target hyperparameters to train the shadow models.
Furthermore, we study the empirical privacy risk of unaccounted use of training data for hyperparameter optimization (HPO) in differentially private (DP) transfer learning. 
We find no statistically significant evidence that performing HPO using training data would increase vulnerability to MIA. 
\end{abstract}

\begin{figure}
    \centering
    \includegraphics[width=\columnwidth]{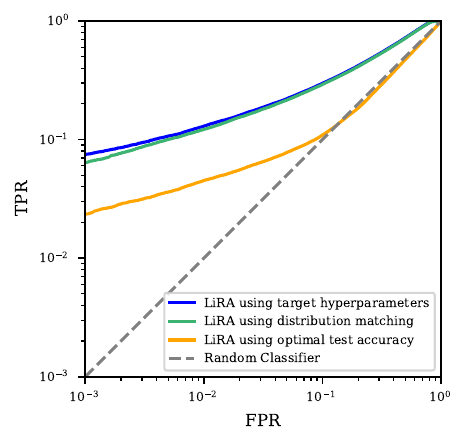}
    \caption{Comparing the performance of attacks using different hyperparameter selection strategies. Selecting the hyperparameters that maximize the similarity between the output distribution of scores of the target model and the shadow model (LiRA using distribution matching) yields an attack as powerful as LiRA when target hyperparameters are used to train the shadow models. On the other hand, training shadow models with the hyperparameters that yield maximum test accuracy on the shadow data sets (LiRA using optimal test accuracy) yields a significantly weaker attack. This shows that selecting hyperparameters for training shadow models is a non-trivial aspect of MIAs.}
    \label{fig:intro_plot}
\end{figure}

\section{Introduction}
Membership inference attacks (MIAs) \cite{shokri_membership_2016} are widely used to empirically test the privacy properties of trained machine learning (ML) models.
These attacks typically involve training so-called \emph{shadow models} whose performance should mimic that of the target model.

As observed from \Cref{fig:intro_plot}, choosing good hyperparameters to train the shadow models is critical to the success of MIAs.
Previous work on MIAs \cite{shokri_membership_2016, carlini_membership_2021, ye_enhanced_2021, zarifzadeh_low-cost_2023, liu_membership_2022} has glossed over the issue of how to choose hyperparameters to train shadow models for MIA, mostly by assuming that the hyperparameters used to train the target model are known to the attacker and can be used in shadow model training.
This assumption is not always valid, as it is often easy for the model creator not to share the hyperparameters. Salem et al. \cite{DBLP:conf/ndss/Salem0HBF019} briefly address performing MIAs in a setting where the target hyperparameters are unknown by showing that their attack is relatively insensitive to hyperparameters. However, they do not contribute an algorithm that could be used to find the optimal hyperparameters for training the shadow models.

In this paper, we explore two aspects of hyperparameters related to MIA.
First, we show that using good hyperparameters when training shadow models for MIA is indeed important.
Not all settings of hyperparameters that yield high utility are good for shadow models, though.
By developing a novel attack with comparable accuracy to known target model hyperparameters, we also show that hiding hyperparameters does not improve privacy.

Second, we apply our attack to study the empirical privacy leakage from hyperparameter adaptation in differentially private deep learning.
Differential privacy (DP) \cite{dwork2006epsilondelta} allows formally proving that the loss of privacy of data subjects from the release of the output of an algorithm is bounded.
DP has been widely adopted to preserve privacy in a broad range of applications, including machine learning (ML) and deep learning.
Algorithms used in deep learning are iterative and access sensitive data multiple times.
One of major strengths of DP is its composition theory: the privacy guarantees weaken predictably under repeated access to sensitive data.

A typical ML workflow would first perform some form of hyperparameter optimisation (HPO) to find good hyperparameters for training the model using a DP stochastic optimisation algorithm such as DP stochastic gradient descent (DP-SGD). 
Although only the final model is typically released, the full privacy bound needs to take into account every access to the sensitive data that led to this result.
A naive privacy bound for HPO would be a composition over every run performed during HPO, which would typically significantly weaken the bounds compared to just a single run.
There are a few methods \cite{liu_private_2019,papernot_hyperparameter_2021,koskela_practical_2023,wang_dp-hypo_2023} that can yield tighter privacy bounds for HPO, but they require using specific HPO protocols (typically random search with a random number of steps), and still yield substantially weaker bounds compared to just a single run.
One recent method \cite{wang_dp-hypo_2023} allows going beyond random search, but at the cost of even weaker bounds.

While it has been shown that the current so-called privacy accounting methods for evaluating the privacy bounds for DP-SGD training can produce tight bounds~\cite{Nasrtightaccountingdp}, limited empirical studies of the tightness of the privacy bounds for HPO have been presented.
There is an example demonstrating privacy leakage through HPO \cite{papernot_hyperparameter_2021}, but this involves specifically releasing the hyperparameters in a non-DP setting. A concurrent work by Xiang et al. \cite{xiang2024revisitingdifferentiallyprivatehyperparameter} audits empirical privacy leakage due to DP-HPO in the less restrictive white-box setting.
Our question focuses on the additional leakage of a released DP model whose hyperparameters have been optimised using training data. In theory, every access to the sensitive training data can increase privacy loss. We test whether this actually leads to higher MIA vulnerability compared to models where HPO has been performed on a different data set from the same population as the training data. Since the both the HPO data and training data are drawn from the same population, the former being a reasonable substitute of the latter for HPO.

\subsection{Our Contributions} 
In this paper, we focus on the transfer learning setting for image classification tasks, where a large pretrained neural network is fine-tuned on a sensitive data set. In our work, we assume that the attacker has access to only the final version of the machine-learning model for MIA. We build on the Likelihood Ratio Attack (LiRA) framework proposed by Carlini et al. \cite{carlini_membership_2021}. 
To summarize, our research offers the following contributions:
\begin{itemize}
    \item We show that score-based MIAs are sensitive to the hyperparameters used for training the shadow models and poor choice of hyperparameters may lead to poor attack performance. 
    
    \item In \Cref{sec:attack_methods}, we propose a novel membership inference attack, \emph{KL-LiRA}, which does not require knowledge of the hyperparameters used to train the target model. KL-LiRA identifies hyperparameters to train the shadow models such that the loss distribution of the shadow models is similar to the loss distribution of the target model.
    
    \item In \Cref{sec:attack_results}, we empirically examine the effectiveness of KL-LiRA and find that it performs nearly as well as an attack that relies on the target model's hyperparameters and is an improvement over any approach that does not identify fitting target model hyperparameters. We use KL-LiRA to recover the hyperparameters for the optimizer used to train the target model which are not known to the attacker.

    \item In \Cref{sec:hpo_leakage} and \ref{sec:hpo_leakage_results}, we examine the impact of using training data for hyperparameter optimization (HPO) on the MIA vulnerability of the final model under differential privacy (DP). We observe no statistically significant increase in MIA vulnerability when HPO is performed on training data compared to using a separate, disjoint dataset for HPO.

\end{itemize}

\section{Background and Preliminaries}
We begin by outlining key aspects of privacy-preserving machine learning relevant to our work.

\subsection{Membership Inference Attacks}

Membership Inference Attacks, or MIAs, are a class of privacy attacks. MIAs are used to infer whether a given sample was used to train a target machine learning model or not. A recent report published by the National Institure of Standards and Technology (NIST) \cite{nist_aml} mentions MIAs as attacks that lead to confidentiality violation by allowing an attacker to determine whether an individual record was included in the training data set of an ML model. 

Generally, the output of MIAs is binary with $0$ indicating that $(x,y)$ was not in the training data set $\data$ of $\model$ whereas $1$ implies that $(x,y) \in \data$. Due to the binary nature of the output, MIAs usually involve training a binary classifier to distinguish whether $(x,y) \in \data$ or $(x,y) \notin \data$. 

\subsubsection{Types of MIA}
There are many different types of MIAs~\cite{hu_membership_2021} that can be broadly categorized based on the knowledge of the attacker. 
An attacker is most powerful when they have White-Box access to the training algorithm. That is, they have knowledge of the full model including the model weights, architecture and intermediate computations in hidden layers which they can exploit to attack a machine learning model.
Another category of MIAs assume the attacker has different degrees of Black-Box access to the training algorithm. Usually, it is assumed that the attacker can only obtain the prediction vector from the model based on input provided by the attacker~\cite{Nasr2019Comprehensive}. Attacks that can be deployed in this setting can be further divided into two major sub-categories: label-only MIAs \cite{choquettechoo2021labelonly} where the attacker can only access the predicted labels when they query the target model, and score-based MIAs \cite{liu_membership_2022, shokri_membership_2016} which follow a general assumption that the attacker has access to the target model's confidence score. In our work, we focus on a class of score-based MIAs that relies on training multiple shadow models to emulate the behaviour of the target model for attack. 

\subsubsection{Measuring MIA accuracy}
The binary classifier used by the attacker in MIA to 
predict whether the sample belongs to the training data or not 
can be used to measure the strength of the MIA. 

For the rest of the paper, we will use the true positive rate ($\tpr$) at a 
specific false positive rate ($\fpr$) for this binary classifier as a measure of the vulnerability. Identifying even a small number of samples with high confidence is considered 
harmful~\cite{carlini_membership_2021} and thus we focus on the regions of small $\fpr$. In some figures, we display the Receiver Operating Characteristic (ROC) which 
plots the tradeoff between $\fpr$s and $\tpr$s.

\subsection{Likelihood Ratio Attack (LiRA)}
Among the different approaches to score-based MIAs \cite{hu_membership_2021}, we use the \textbf{Likelihood Ratio Attack} (LiRA)~\cite{carlini_membership_2021} in our work. Other score-based MIAs, such as RMIA~\cite{zarifzadeh_low-cost_2023}, have been proposed as an improvement over LiRA when the attacker cannot afford to train a large number of shadow models for MIA. RMIA is shown to match the performance of LiRA when it is possible to train many shadow models for the attack. Given that our experiments require training large number of shadow models, we use LiRA as the representative score-based attack in our work.

In LiRA, the attacker trains multiple shadow models using the information available about the target model. Given that all ML models are trained to minimize their loss on the training samples, LiRA exploits this fact by training multiple shadow models such that for 1/2 of them, the target sample $(x,y)$ is IN the training data set ($x \in \data$), while for the other half, $(x,y)$ is OUT of the training data set ($x \notin \data$).

In the LiRA paper~\cite{carlini_membership_2021}, the authors apply logit scaling to the model's ($\model$) predicted confidence score for a given sample, $(x,y)$, to approximate the model's output using a Normal distribution,
\begin{equation}
\label{eq:logits}
    \logits(p) = \log \left(\dfrac{p}{1-p}\right), \text{ where } p = \model(x)_y.
\end{equation}

Using the predicted (and logit-scaled) confidence scores of the shadow models on the target sample, the attacker builds IN and OUT Gaussian distributions. Finally, the attacker uses a likelihood ratio test on these distributions to determine whether $(x,y) $ was used for training the target model. In our experiments, we use an optimized version of LiRA proposed by Carlini et al. \cite{carlini_membership_2021} as the baseline wherein the attacker uses the target model's hyperparameters to train the shadow models for MIA.

\subsection{Differential Privacy (DP)}
Differential privacy (DP)~\cite{dwork2006epsilondelta} is a framework for protecting the privacy of sensitive data used for data analysis and provides provable guarantees.
The commonly used version of DP called $(\varepsilon, \delta)$-DP quantifies the privacy loss using a privacy budget consisting of $\varepsilon \geq 0$ and $\delta \in [0, 1]$, where smaller values correspond to a stronger privacy guarantee.

\subsubsection{Deep Learning under DP}
DP-SGD~\cite{pmlr-v22-rajkumar12, Song2013StochasticGD, abadi_deep_2016} is a modification of the stochastic gradient descent (SGD) algorithm that guarantees DP in deep learning. In every step, DP-SGD computes per-example gradients, clips them and then adds noise to the aggregated gradient. A privacy accountant is used to quantify the privacy budget. Training models under DP introduce a privacy--utility trade-off. A smaller privacy budget requires adding more noise, but this degrades the utility. Furthermore, using DP introduces additional parameters that are specific to privacy, such as the gradient norm clipping bound and the amount of noise which require careful tuning during HPO.
We would like to refer to a comprehensive guide on training ML models under DP~\cite{Ponomareva2023DP-fy}.

\subsubsection{High utility models under DP}
Training high utility models under DP from scratch is challenging. There is a significant gap between training under DP and without DP even for simple computer vision benchmarks~\cite{tramer_considerations_2022}, resulting in models trained from scratch under DP being unsuitable for real world deployments.
The current state-of-the-art results are obtained through transfer learning~\cite{yosinski2014transferable} under the assumption that a public non-sensitive data set can be utilized for pre-training and only the fine-tuning data set is sensitive and needs to be protected through DP.
Prior work has shown that transfer learning under DP is effective for both vision~\cite{cattan2022fine,kurakin2022toward,tobaben_efficacy_2023,tito2024privacyaware} and language tasks~\cite{li2022large,yu2022differentially}. 
Parameter-efficient fine-tuning using adapters like LoRA~\cite{hu_lora_2021} and FiLM~\cite{perez_film_2018} are competitive in comparison to fine-tuning all parameters under DP as they yield a similar privacy-utility trade-off at a much smaller computational cost~\cite{yu2022differentially,tobaben_efficacy_2023}. 

\begin{table}[t]
    
    \caption{A Summary of the notations used in the paper.}
    \label{tab:terms}
    \renewcommand{\arraystretch}{1.2}
    \centering
    \adjustbox{max width=\columnwidth}{\begin{tabular}{l|r}
    \hline
         \textbf{Notation} & \textbf{Description}  \\
    \hline
    $\mathbb{D}$ & Data generating distribution \\
    $\model$ & Model obtained after running a training algorithm \\
    $(x,y)$ & Target sample s.t. $(x,y) \sim \mathbb{D} $ \\
    $\data_{i}$ & Shadow training data of $i$th shadow model $\model_{i}$, s.t. $\data_{i} \sim \mathbb{D}$ \\
    $\trainfunction()$ & Training function \\
    $\hpofunction()$ &  Function for Hyperparameter Optimization \\
    $\eta_{i}$ & Hyperparameters, $\eta_{i} = \text{} \hpofunction$ ($\data_i$) \\
    $\model_{\data_i,\eta_j}$ & Shadow model, $\model_{\data_i,\eta_j} = \text{} \trainfunction(\data_i, \eta_j)$ \\
    $\mathcal{N}(\mu,\sigma^2)$ & Gaussian distribution with mean $\mu$ and variance $\sigma^2$ \\
    $B(n,p)$ & Binomial distribution with $n$ trials and $p$ probability of success \\
    $M$ & Number of shadow models \\
    $S$ & Examples per class for few-shot training \\
    $\prob$ & Probability distribution \\
     \hline
    \end{tabular}}
\end{table}

\subsubsection{HPO under DP}
From the strict theory perspective of DP, performing HPO on the training set requires accounting for the privacy loss originating from the HPO~\cite{koskela_practical_2023, papernot_hyperparameter_2021, liu_private_2019, wang_dp-hypo_2023}.
Prior work that achieves state-of-the-art results usually ignores the privacy cost of performing HPO on the training set~\cite{li2022large,yu2022differentially,cattan2022fine,de2022unlocking,kurakin2022toward,tobaben_efficacy_2023}. 
The existing works on differentially private hyperparameter optimization (DP-HPO)
\cite{koskela_practical_2023, papernot_hyperparameter_2021, liu_private_2019, wang_dp-hypo_2023} account for the privacy leakage through HPO assuming that the trained model and the corresponding set of hyperparameters are released under DP, assuming a specific HPO protocol.
When performing HPO on the training set using a standard HPO algorithm with DP training, the best known privacy bounds for releasing the final model would require evaluating a composition over every model trained during HPO, even if only the final model is released.
The resulting DP bounds will be substantially weaker than bounds from single model training.

\subsubsection{Relationship between MIA and DP}
Empirical lower bounds on the privacy leakage obtained through MIAs can complement the theoretical DP upper bounds. Consequently, they have been employed for empirically determining the privacy of training data under varying threat models\cite{DBLP:journals/corr/abs-2101-04535, NEURIPS2020_fc4ddc15}. Any classifier distinguishing the training samples based on the results of a DP algorithm has an upper bound for the $\tpr$ that depends on the privacy parameters $(\varepsilon, \delta)$~\cite{Kairouz2015Composition}. Since MIAs are trying to build exactly these types of classifiers, we can use the upper bounds in \Cref{thm:Kairouz_bound} to validate the privacy claims, and also to better understand the gap between the theoretical privacy guarantees of DP and empirical results obtained through MIA.

\begin{theorem}[\cite{Kairouz2015Composition}]
A mechanism $\mathcal{M}: \mathcal{X} \to \mathcal{Y}$ is $(\varepsilon, \delta)$-DP if and only if for all adjacent $\mathcal{D}\sim\mathcal{D}'$, every test for distinguishing $\mathcal{D}$ and $\mathcal{D}'$ satisfies

\begin{equation}
\label{eq:tpr_bound}
    \tpr \leq \min\{ e^\varepsilon \fpr + \delta,
    1 - e^{-\varepsilon} (1 - \delta - \fpr) \} \ .
\end{equation}
\label{thm:Kairouz_bound}
\end{theorem}

\Cref{tab:terms} provides a comprehensive summary of all terms and notations used throughout the remainder of the paper.

\section{Attacks: Methods}
\label{sec:attack_methods}

To build the shadow models for LiRA, the attacker typically
needs to have access to some additional information about the training
of the target model. For majority of their experiments in the LiRA paper, Carlini et al.
\cite{carlini_membership_2021}
assume that the attacker has access to the target model's 
architecture $\mathcal{A}$, the hyperparameters $\eta$
and distribution of the training data $\mathbb{D}$, although they do briefly study the attacks with mismatched training procedures for the target and shadow models. For the rest of
the paper, we assume that the attacker has access to $\mathbb{D}$,
but might not have access to the correct architecture or 
exact hyperparameters for the optimizer used to train the target model. However, the attacker does know what constitutes a hyperparameter for a given target model and the corresponding range of values as detailed in \Cref{tab:hyperparam_ranges} in the Appendix.
We use $\attack(\textbf{aux})$
to denote the LiRA attack with different levels of access to the auxiliary information.
Specifically, we will focus on three threat models: 
\begin{inlinelist}
    \item $\attack(\mathcal{A}, \eta)$ where 
    the attacker has access to both the target model's architecture
    and hyperparameters,
    \item $\attack(\mathcal{A})$ with access to the 
    target architecture but not to the hyperparameters, and
    \item $\bb$ with access to neither the architecture
    nor the hyperparameters.
\end{inlinelist}
The threat models and the corresponding auxiliary information is 
summarized in \cref{tab:settings_and_attacks}.
A generalized LiRA algorithm is provided in \Cref{alg:lira}. The algorithm accepts as its input a set of hyperparameter values (highlighted in \textcolor{blue}{blue}) that are used to train the shadow models for the attack. 

\begin{table}[htbp]
    \footnotesize
    \caption{Threat Models and the possible versions of LiRA that can be employed under each setting.}
    \label{tab:settings_and_attacks}
    \centering
    \renewcommand{\arraystretch}{1.3}
    \begin{tabular}{ccccccc}
    \hline
    \multicolumn{1}{c}{\multirow{2}{*}{Threat Model}} & \multicolumn{4}{c}{Auxiliary Information} &  \multicolumn{1}{c}{\multirow{2}{*}{Algorithms}}\\
    \multicolumn{1}{c}{} & $\mathbb{D}$ & $\mathcal{A}_{\mathcal{T}}$ & $\eta_{\mathcal{T}}$ & Model Weights & \multicolumn{1}{c}{} \\
    \hline
    $\attack(\mathcal{A}, \eta)$ & \checkmark & \checkmark & \checkmark & - &  LiRA\\
    \hline
    \multirow{2}{*}{$\attack(\mathcal{A})$} & \multirow{2}{*}{\checkmark} & \multirow{2}{*}{\checkmark} & \multirow{2}{*}{-} & \multirow{2}{*}{-} &  ACC-LiRA \\
     &  &  &  &  &  KL-LiRA \\
    \hline
    \multirow{2}{*}{$\bb$} & \multirow{2}{*}{\checkmark} & \multirow{2}{*}{-} & \multirow{2}{*}{-} & \multirow{2}{*}{-} &  ACC-LiRA \\
     &  &  &  &  &  KL-LiRA \\
    \hline
    \end{tabular}
\end{table}

\subsection{Strategies for Selecting Shadow Model Training Hyperparameters}
\label{subsec:IIIA}
When the target hyperparameters $\eta_\mathcal{T}$ are made available together with the
target model, the attacker can use the them for 
training the shadow models as done by Carlini et al. \cite{carlini_membership_2021} for LiRA.To implement LiRA, the inputs to \Cref{alg:lira} will be $\{\eta_\mathcal{T}\}_{j=1}^{M}$.

However, when $\eta_\mathcal{T}$ is not available, it is not clear how to select the hyperparameters that would be optimal to train the shadow models. 
The hyperparameters used for training can make a big difference in the performance and behaviour of a model. Therefore, to simulate the behaviour of the target model in the shadow models, the hyperparameters for the shadow models need to be also carefully selected. Next, we will discuss the two approaches for HPO for 
shadow model training.

\subsubsection{Accuracy-based HPO for shadow models (ACC-LiRA)}
\begin{figure*}[!htp]
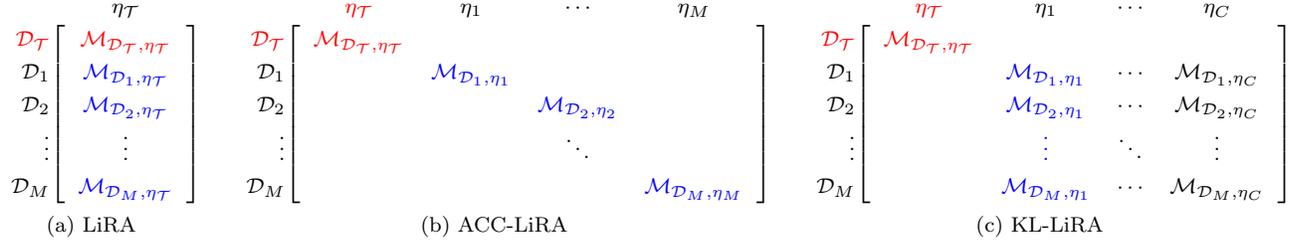

    \footnotesize
    \centering
    \subfloat[LiRA]{
    \label{subfig:LiRA}
     \renewcommand{\arraystretch}{1.3}
    {\NiceMatrixOptions{columns-width=0.1mm}
    $\begin{bNiceArray}{c}[first-row,first-col,nullify-dots,margin,rules/color=blue]       
    & \eta_{\mathcal{T}} \\
     \textcolor{red}{\data_{\mathcal{T}}} & \textcolor{red}{\model_{\data_{\mathcal{T}},\eta_{\mathcal{T}}}} \\
    \data_1 & \textcolor{blue}{\model_{\data_1,\eta_{\mathcal{T}}}} \\
     \data_2 & \textcolor{blue}{\model_{\data_2,\eta_{\mathcal{T}}}} \\
     \vdots & \vdots \\
       \data_M & \textcolor{blue}{\model_{\data_M,\eta_{\mathcal{T}}}} \\
    \end{bNiceArray}$}}
    ~
    \subfloat[ACC-LiRA]{
    \label{subfig:ACC-LiRA}
    \renewcommand{\arraystretch}{1.3}
    {\NiceMatrixOptions{columns-width=0.05mm}
    $\begin{bNiceArray}{cccc}[first-row,first-col,nullify-dots,margin,rules/color=blue]       
    & \textcolor{red}{\eta_{\mathcal{T}}} & \eta_1 & \cdots  & \eta_M \\
    \textcolor{red}{\data_{\mathcal{T}}} & \textcolor{red}{\model_{\data_{\mathcal{T}},\eta_{\mathcal{T}}}} \\
    \data_1 & & \textcolor{blue}{\model_{\data_1,\eta_1}} \\
     \data_2 &  &  & \textcolor{blue}{\model_{\data_2,\eta_2}} \\
     \vdots &  &   &  \ddots \\
     
     \data_M &  &  &   &  \textcolor{blue}{\model_{\data_M,\eta_M}}\\
    \end{bNiceArray}$}}
    ~
    \subfloat[KL-LiRA]{
    \label{subfig:KL-LiRA}
    \renewcommand{\arraystretch}{1.3}
    {\NiceMatrixOptions{columns-width=0.1mm}
    $\begin{bNiceArray}{cccc}[first-row,first-col,nullify-dots,margin,rules/color=blue]       
    & \textcolor{red}{\eta_{\mathcal{T}}} & \eta_1 & \cdots  & \eta_C   \\
    \textcolor{red}{\data_{\mathcal{T}}} & \textcolor{red}{\model_{\data_{\mathcal{T}},\eta_{\mathcal{T}}}} &  &    &  \\
    \data_1 &  &  \textcolor{blue}{\model_{\data_1,\eta_1}} & \cdots   & \model_{\data_1,\eta_C} \\
    \data_2  &  &  \textcolor{blue}{\model_{\data_2,\eta_1}}  &  \cdots  & \model_{\data_2,\eta_C} \\
    \vdots &  & \textcolor{blue}{\vdots}  &  \ddots  & \vdots \\
    \data_M  &  & \textcolor{blue}{\model_{\data_M,\eta_1}}  & \cdots    & \model_{\data_M,\eta_C} \\
    \end{bNiceArray}$}}
    \caption{The schematic illustration of the different attacks as described in \Cref{sec:attack_methods}. $\model_{\data_i,\eta_j}$  represents a model trained with $\data_i$ data set and $\eta_j$ hyperparameters. The target model and corresponding data set/ hyperparameters are highlighted in \textcolor{red}{red} to denote that the attacker has restricted access to them whereas the corresponding shadow models are highlighted in \textcolor{blue}{blue}.}
    \label{fig:schema_for_attacks}
\end{figure*}

Since the attacker does not have the access to the training data
of the target model, the attacker cannot simply replicate the HPO
that was originally executed to obtain the hyperparameters used to train
the target model. However, it is possible for the attacker to use the shadow data
sets for HPO. Given a set of shadow data sets 
$\{\data_i\}_{i=1}^M$, the attacker performs HPO on
each $\data_i$ and obtains a set of $M$ hyperparameters $\{\eta_i\}_{i=1}^{M}$:
\begin{align}
    \eta_i = \textsc{HPO}(\data_i).
\end{align}
The attacker can then proceed to use $\{\eta_i\}_{i=1}^{M}$ as an input to \Cref{alg:lira} to train the shadow models. \Cref{subfig:ACC-LiRA} depicts the hyperparameter optimization for a target model $\model_{\data_{\mathcal{T}},\eta_{\mathcal{T}}}$ using the shadow data sets $\{\data_{i}\}_{i=1}^{M}$ for the HPO and shadow model training.
We call this setting the \emph{accuracy-based LiRA}
or \textbf{ACC-LiRA}. 

This procedure guarantees that the shadow models are trained with hyperparameters that optimize their utility. In LiRA, the shadow models are intended to simulate the behaviour of the target model as closely as possible. Therefore, assuming that the target model's hyperparameters are tuned with respect to similar utility targets, the ACC-LiRA presents a reasonable approach for training LiRA without additional information about the
hyperparameters. 

\begin{algorithm}[t]
\caption{\textbf{LiRA:} Generalized Likelihood Ratio Attack (LiRA) per Carlini et al. \cite{carlini_membership_2021}. Given shadow data sets $\{\data_i \}_{i=1}^M$, \textcolor{blue}{a set of hyperparameters  $\{\eta_i\}_{i=1}^M$}, model architecture $\architecture$, the attacker aims to use the hyperparameters to train $\numbershadowmodels$ shadow models for each of the shadow data sets. Differences with standard LiRA are highlighted in blue.}
\label{alg:lira}
\begin{algorithmic}[1]
\REQUIRE 
target model $\model_{\mathcal{T}}$, 
target sample $(x,y)$, 
\textcolor{blue}{$\{\eta_i\}_{i=1}^M$}, 
$\{\data_i\}_{i=1}^M$, $\architecture$
\STATE Set $\text{confs}_{\text{in}} \leftarrow \{ \}$,  $\text{confs}_{\text{out}} \leftarrow \{ \}$
\FOR {$i = 1,...,\numbershadowmodels$}
\STATE $\model_{\text{in}} \leftarrow \trainfunction( \architecture, \data_i \cup \{(x,y)\},\textcolor{blue}{\eta_i})$ 
\STATE $\text{confs}_{\text{in}} \leftarrow \text{confs}_{\text{in}} \cup \{\logits(\model_{\text{in}}(x)_y)\}$
\STATE $\model_{\text{out}} \leftarrow \trainfunction(\architecture, \data_i \setminus \{(x,y)\},\textcolor{blue}{\eta_i})$ 
\STATE $\text{confs}_{\text{out}} \leftarrow \text{confs}_{\text{out}} \cup \{\logits(\model_{\text{out}}(x)_y)\}$
\ENDFOR
\STATE $\mu_{\text{in}} \leftarrow \texttt{mean}(\text{confs}_{\text{in}})$
\STATE $\mu_{\text{out}} \leftarrow \texttt{mean}(\text{confs}_{\text{out}})$
\STATE $\sigma^2_{\text{in}} \leftarrow \texttt{var}(\text{confs}_{\text{in}})$
\STATE $\sigma^2_{\text{out}} \leftarrow \texttt{var}(\text{confs}_{\text{out}})$
\STATE $\text{conf}_{\model_{\mathcal{T}}} \leftarrow \logits(\model_{\mathcal{T}}(x)_y)$ 
\RETURN Membership Score $\leftarrow \dfrac{\prob(\text{conf}_{\model_{\mathcal{T}}}| \normal(
\mu_{\text{in}},\sigma^2_{\text{in}}))}{\prob(\text{conf}_{\model_{\mathcal{T}}}| \normal(\mu_{\text{out}},\sigma^2_{\text{out}}))}$
\end{algorithmic}
\end{algorithm}

\begin{algorithm}[t]

\caption{\textbf{Hyperparameter Selection in KL-LiRA:} Given a set of shadow data
sets $\{ \data_i \}_{i=1}^N$, a set of candidate hyperparameters $\{ \eta_j \}_{j=1}^C$, model architecture $\architecture$ and a target model $\modeltar$, this algorithm aims to find the optimal hyperparameter $\eta_{j^*}$, for which the loss distribution over the shadow models is most similar to the loss distribution of the target model on the shadow datasets.}
\label{alg:kl-lira}
\begin{algorithmic}[1]
\REQUIRE 
$\{ \data_i \}_{i=1}^N$, $\{ \eta_j \}_{j=1}^C$, $\modeltar$, $\architecture$
\STATE 
Set $\Bar{\phi} \leftarrow \{ \}$
\FOR {$j = 1,...,C$}
\STATE $\phi_j \leftarrow \{ \}$
\FOR{$i = 1,...,N$}
\STATE 
    $\mathcal{M}_{\mathcal{S}} \leftarrow \trainfunction(\architecture, \data_i, \eta_j)$
\STATE 
    $\normal_{\mathcal{T}} \leftarrow \textsc{normal-approx}(\modeltar(\data_i))$
\STATE 
    $\normal_{\mathcal{S}} \leftarrow \textsc{normal-approx}(\modelshadow (\data_i))$
\STATE 
    $\phi_{j} \leftarrow \phi_{j} \cup 
        \{\phi_{KL}(\normal_{\mathcal{T}} || \normal_{\mathcal{S}}) \}$ 
\ENDFOR
\STATE $\Bar{\phi} \leftarrow \Bar{\phi} \cup \{\texttt{mean}({\phi_j})\}$
\ENDFOR
\STATE  $j^* \leftarrow \texttt{argmin}(\Bar{\phi})$ 
\RETURN $\eta_{j^*}$
\end{algorithmic}
\end{algorithm}

\subsubsection{Distribution-based HPO for shadow models (KL-LiRA)}
However, it is not clear whether the optimal hyperparameters with respect to some utility criterion make the shadow models behave similarly to the target model. It could be that while both the shadow and target models provide good aggregate utility 
in some task, the two models still lead to significantly different
outcomes on a single sample.

Recall that the attacker can query the target model and has
access to a collection of data sets or a data distribution similar
to the target data set. Hence, the attacker can build a distribution
of the scores corresponding to the data points by passing
the shadow data points through the target model. Now, this 
distribution characterizes the behaviour of the target models 
outcomes. Therefore, if we can make sure that the shadow
models behave similarly to the target by optimizing the
hyperparameters in a way that makes the shadow models' output
distribution similar to that of the target model.

This leads to our final attack, the Kullback--Leibler LiRA or 
\textbf{KL-LiRA}. In KL-LiRA, we use the Kullback--Leibler (KL)
divergence, a common measure of the distance between
two probability distributions, to compute the similarity between
loss distributions from the target model and a shadow model on the
corresponding shadow data set. 
Carlini et al.\cite{carlini_membership_2021} showed that the
loss distributions for LiRA can be well approximated with a
Gaussian. Therefore, we approximate both the target and shadow
loss distributions as Gaussians and choose the shadow 
hyperparameters that minimize:
\begin{equation}
    \label{eq:kld}
    \phi_{\text{KL}}(\normal_{\mathcal{T}}||\normal_{\mathcal{S}}) = \dfrac{1}{2}\Big[ \dfrac{(\mu_{\mathcal{S}}-\mu_{\mathcal{T}})^2}{\sigma_{\mathcal{S}}^2} + \dfrac{\sigma_{\mathcal{T}}^2}{\sigma_{\mathcal{S}}^2} - \text{ln} \dfrac{\sigma_\mathcal{T}^2}{\sigma_{\mathcal{S}}^2} - 1 \Big],
\end{equation}
where both the shadow ($\mathcal{S}$) and target ($\mathcal{T}$) 
distributions' means and variances $\mu, \sigma^2$ are estimated from the sampled losses. 
The pseudocode for the hyperparameter selection of 
KL-LiRA is illustrated in \Cref{alg:kl-lira}. Once we get the set of optimal hyperparameters for KL-LiRA, we can forward them as inputs ($\{\eta_{j^*}\}_{i=1}^M$) to \Cref{alg:lira} to run the attack.

Running the KL-LiRA hyperparameter selection requires a set of the candidate hyperparameters $\{ \eta_j\}_{j=1}^C$. 
This set should reasonably reflect the attacker's prior beliefs on which hyperparameters the target model was using.  
One approach for obtaining the set is to perform HPO on a subset of the available shadow data selected at random. We select $C$ of the available $M$ shadow data sets for HPO and use the resulting hyperparameters as the candidate set, i.e. build $\{\eta_j\}_{j=1}^C$ with,
\begin{align}
    \eta_j = \texttt{HPO}(\data_j)
\end{align}
where $\data_j$ is drawn from training data distribution. This procedure for a target model $\model_{\data_{\mathcal{T}},\eta_{\mathcal{T}}}$ is depicted in \Cref{subfig:KL-LiRA}.

\subsubsection{Summary of the attacks}
Next, we will summarize the three attacks we consider in this paper. Each attack will call \cref{alg:lira} with the set hyperparameters for the shadow models outlined in the following:
\begin{description}
    \item[\textbf{LiRA}] 
    \begin{align}
        \{\eta_i\}_{i=1}^M = \{\eta_{\mathcal{T}}\}_{i=1}^M,
    \end{align}
    where $\eta_\mathcal{T}$ are the target model's hyperparameters.
    \item[\textbf{ACC-LiRA}] 
    \begin{align}
        \{\eta_i\}_{i=1}^M = \{\texttt{HPO}(\data_i)\}_{i=1}^M.
    \end{align}
    \item[\textbf{KL-LiRA}]
    Attacker obtains a set of candidate hyperparameters from shadow data sets
        \begin{align}
            \{\eta_j\}_{j=1}^C = \{\texttt{HPO}(\data_j)\}_{j=1}^C.
        \end{align}
    Finds the optimal $\eta_{j^*}$ using \cref{alg:kl-lira}
    and sets
        \begin{align}
            \{\eta_i\}_{i=1}^M = \{\eta_{j^*}\}_{i=1}^M.
        \end{align}
\end{description}

\subsection{Computational Cost of Different Attacks}
\label{subsec:compcostattacks}
The computational cost of running an attack can be expressed in terms of the number of shadow models the attacker will need to train for the attack, as well as the number of inference queries on the target model.

The cost in terms of inference queries is the same for all variants of LiRA. This cost is equal to the number of samples in the union of the shadow data sets $\cup_{i=1}^N \data_i$. For KL-LiRA, the inference queries on Line $6$ of \Cref{alg:kl-lira} can be answered by using cached responses to these queries.

LiRA attack requires no additional computation to run HPO since the target hyperparameters are known to the attacker. Thus its total number of models needed to train is $M$.

ACC-LiRA incurs an additional cost for attacker since they will have to run HPO for each of the $\numbershadowmodels$ shadow models. Let $T$ be the number of HPO trials. Then, the total number of models needed to train is $\numbershadowmodels \times T$ which is $T$-fold increase compared to LiRA if $\numbershadowmodels$ is regarded as a constant.

For KL-LiRA, the attacker begins by randomly selected $C$ data sets from $\{\data_i\}_{i=1}^{M}$ and runs HPO on each of these to obtain candidate hyperparameters $\{\eta_j \}_{j=1}^C$. The number of models needed to train for HPO under KL-LiRA will thus be $C \times T$. In the next step, the attacker proceeds to train $N$ models for each of the $\eta_j$ using $N$ data sets selected at random from the $M$ shadow data sets. These models are used by the attacker for hyperparameter selection using \Cref{alg:kl-lira}. Since the attacker has already trained $N$ models on shadow data sets, to run the attack with $M$ shadow models, they will need to train models for the remaining $M-N$ shadow data sets.
This brings the total number of models needed to train for KL-LiRA to $ (C\times T) + C\times (N-1) + M - N$. 

\subsection{Testing MIA Against a Training Algorithm}
\label{subsec:miagrid}

\Cref{alg:lira} is defined to attack a single model and a single target sample. However, if MIA is deployed for testing the privacy of the training algorithm and not just a specific sample or model, we will be required to train multiple models using to estimate the MIA vulnerability over the algorithm. This would be computationally inefficient if the attacker plans to use naive ACC-LiRA or KL-LiRA since the computational cost of running such an attack using \Cref{alg:lira} would scale with the number of target models. Furthermore, for each target model,  using ACC-LiRA or KL-LiRA as an attack would require additional computation for HPO (and hyperparameter selection in case of KL-LiRA) as discussed in the previous section. 

To optimally run MIA against an algorithm, Carlini et al.~\cite{carlini_membership_2021} proposed an efficient implementation of LiRA. 
It involves sampling $M+1$ data sets $\data_0, \data_1, \ldots \data_M$ from the training data set such that the probability of a sample to be selected to each data set is $0.5$, training models for each of these data sets, and finally attacking each of these models while using the others as shadow models.
We can use the same procedure for ACC-LiRA, but not KL-LiRA.

For KL-LiRA, we first perform HPO on each of these data sets which us a set of $M+1$ hyperparameters $\{\eta_j\}_{j=0}^M$ and train the corresponding models to be used as target models. We then randomly select $C+1$ of these hyperparameters to be used as seeds for KL-LiRA hyperparameter selection, and train $N$ models for each of these. For each target model, we select $C$ hyperparameters not including those of the target model, apply \Cref{alg:kl-lira} to find the optimal hyperparameters against that target model, and train the remaining shadow models for those hyperparameters if they have not been trained previously.

The models used in all the attacks are of defined by a pair $(\data_i, \eta_j)$. These are subsets of a $(M+1) \times (M+1)$ matrix of models we call the \emph{MIA-Grid}. The MIA-Grid could be deployed to run any of the aforementioned attacks over multiple target models and data points making it feasible to assess the MIA vulnerability of the training algorithm. 

This approach is not without its disadvantages. By favoring a more efficient implementation of attacks using the MIA-Grid, we end up using shadow data sets that differ by more than one sample to compute the IN and OUT distributions (as opposed to when the data sets differ only in terms of the presence or absence of the target sample as in \Cref{alg:lira}).
This would lead to a weaker attack against individual samples when $\numbershadowmodels$ is limited but the trade-off is acceptable since the MIA-Grid allows us to evaluate MIA vulnerability over the training algorithm and not just one model and/or one data point.

\begin{figure*}[t]
    \centering
    \subfloat[CIFAR10, ViT-B + Head, $S = 100$]{

    \includegraphics[width=0.3\textwidth]{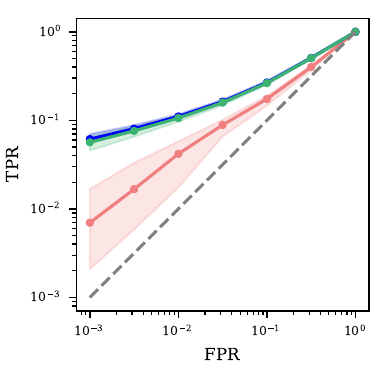}}
    ~
    \subfloat[CIFAR10, ViT-B + Head, $S = 50$]{

    \includegraphics[width=0.3\textwidth]{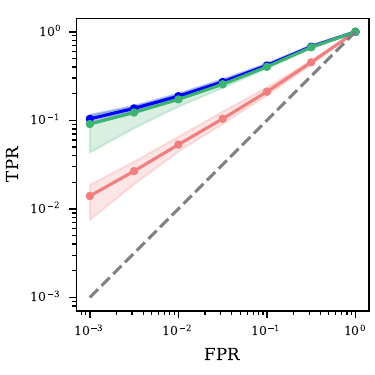}}
    \subfloat[CIFAR100, ViT-B + Head, $S=100$]{

    \includegraphics[width=0.3\textwidth]{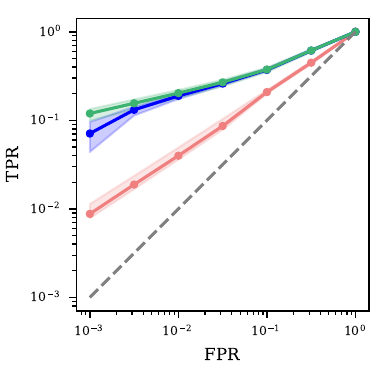}}
    \\
    \subfloat[CIFAR10, R-50 + Head, $S = 50$]{

    \includegraphics[width=0.3\textwidth]{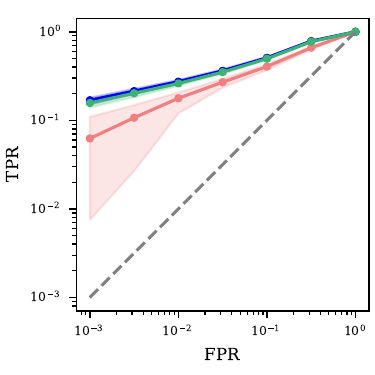}}
    \subfloat[CIFAR10, R-50 + FiLM, $S=50$]{

    \includegraphics[width=0.3\textwidth]{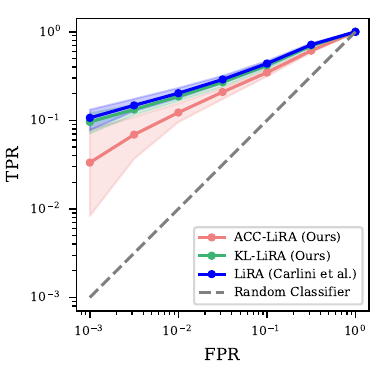}}
    \caption{Overview of attack performances against non-private models. ROC curves depict the performance of different attacks for individual experiments computed over 10 runs. All the plots are generated using 128 shadow models except for FiLM which uses 64 shadow models. The error bars represent the Clopper--Pearson confidence interval associated with the estimated $\tpr$ at fixed $\fpr$. LiRA represents a weaker threat model since the attacker has knowledge of the target hyperparameters. MIA vulnerability measured using KL-LiRA matches the vulnerability measured using LiRA. ACC-LiRA's performs poorly when compared to LiRA and KL-LiRA.}
    \label{fig:roc_plots_mia}
\end{figure*}

\begin{figure}
    \centering
    \includegraphics[width=0.5\textwidth]{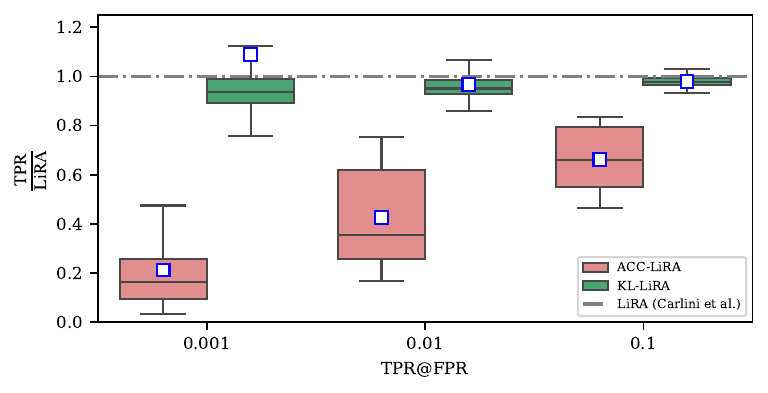}
    \caption{Overview of relative attack performances against non-private models. MIA vulnerability of KL-LiRA and ACC-LiRA relative to LiRA computed over all the experiments. We normalize the $\tpr$ at fixed $\fpr$ for KL-LiRA and ACC-LiRA by the corresponding $\tpr$ for LiRA for each experiment and run.}
    \label{fig:summary_plot_1}
\end{figure}

\section{Experimental Setup}
In this section, we describe the common setup for the experiments. Our experiments focus on models trained with few-shot transfer learning with or without DP.

\subsection{Model Training}
\subsubsection{Fine-Tuning Data Sets} In our experiments, we fine-tuned the models on the CIFAR-10 and CIFAR-100~\cite{krizhevsky2009learning} data sets, which are widely used as benchmark data sets for image classification tasks ~\cite{de2022unlocking,tobaben_efficacy_2023}.

\subsubsection{Few-Shot Training} We fine-tuned our models for the experiments using $S=100$ and $S=50$ shots. We use Adam \cite{kingma_adam_2014} as the optimizer for training the models.

\subsubsection{Pretrained Model Architectures} We fine-tuned both ResNet and Vision Transformer architectures pretrained on ImageNet-21K~\cite{russakovsky_imagenet_2014}:
\begin{inlinelist}
    \item BiT-M-R50x1 (R-50)~\cite{kolesnikov_big_2019} with $23.5$M parameters
    \item and Vision Transformer ViT-Base-16 (ViT-B) \cite{dosovitskiy_image_2020} with $85.8$M parameters. We briefly discuss the attacking models with relatively weaker pretrained architectures in \Cref{appendix:weaker_pretrained_backbone}.
\end{inlinelist}

\subsubsection{Parameterizations} Due to the computational costs of fine-tuning all parameters of the pretrained model, we restricted the fine-tuning to subsets of all feature extractor parameters. These are:
\begin{itemize}
    \item  \textbf{Head:} The head (last layer) of the pretrained model is replaced by a trainable linear layer while all the remaining parameters of the body are kept frozen.
    \item \textbf{FiLM:} In this configuration, along with training the linear layer, we fine-tuned the parameter-efficient FiLM \cite{perez_film_2018} adapters scattered throughout the network. Although there are many other such adapters, such as Model Patch \cite{mudrakarta_k_2018}, LoRA \cite{hu_lora_2021}, CaSE \cite{patacchiola_contextual_2022} etc., we chose FiLM as it has proven to be highly effective in previous work on parameter-efficient few-shot transfer learning \cite{shysheya_fit_2022, tobaben_efficacy_2023}.
\end{itemize}
        
\subsubsection{Hyperparameter Optimization (HPO)} Our HPO protocol closely follows the one used by Tobaben et al. \cite{tobaben_efficacy_2023} since it has been proven to yield SOTA results for (DP) few-shot models with minor differences. We fix the number of epochs for HPO to $40$. In the non-DP setting, we only tune the batch size and the learning rate. When training the models with DP, an additional hyperparameter, the gradient clipping bound, is added to the HPO process.
Details of the HPO process used in the experiments are available in \Cref{appendix1}.

\subsubsection{DP-Adam} For implementing differentially private deep learning, we used the Opacus library~\cite{yousefpour_opacus_2021} that implements DP-Adam on top of PyTorch~\cite{pytorch}. We use the PRV accountant~\cite{Gopi2021prv} to calculate the spent privacy budget.

\subsection{MIA}
We attack the fine-tuned models using LiRA.
\subsubsection{Metrics}  For all experiments, we report the metrics computed over 10 repeats of the attack algorithm. For each repeat, we run the attack algorithm with a new set of shadow data sets.
\begin{itemize}
    \item \textbf{TPR at Low FPR:} We summarize the success rate of MIA as measured by the True Positive Rate ($\tpr$) in the low-\textit{False Positive Rate} ($\fpr$) regime, as recommended by Carlini et al. \cite{carlini_membership_2021}.
    \item \textbf{Receiver Operator Characteristic (ROC):} Additionally, we use the ROC curve (comparing $\tpr$ against $\fpr$) on the log-log scale to visualize the performance of different attacks.
    \item \textbf{Clopper--Pearson Confidence Interval:} For certain plots, we estimate the uncertainty associated with the $\tpr$ over different repeats using the Clopper--Pearson confidence interval \cite{10.1093/biomet/26.4.404}. For $\textsc{tp}$ true positives predicted out of $\textsc{p}$ positives, the $1-\alpha$ confidence interval for the $\tpr$ is given by:
    \begin{equation}
        \begin{split}
            B(\alpha/2; \textsc{tp}, \textsc{p}-\textsc{tp}+1) < \tpr \\ < B(1 - \alpha/2; \textsc{tp}+1, \textsc{p}-\textsc{tp}) 
        \end{split}   
    \end{equation}
\end{itemize}

\subsubsection{LiRA} For LiRA, we maintain $M=128$ shadow models throughout the experiments to build the MIA-Grid except for the training models with FiLM, wherein we reduced $M$ to $64$ owing to the computational constraints of building the MIA-Grid. 
Unlike Carlini et al.\ \cite{carlini_membership_2021}, we do not use train-time data augmentation for the attacks, because these are not as important in fine-tuning. Since we are using $\numbershadowmodels \ge 64$ in our experiments, we estimate the per-example variance when running attacks per the recommendations of Carlini et al. \cite{carlini_membership_2021}.

\section{Attacks: Results}

\label{sec:attack_results}
In this section, we evaluate the performance of different attacks under various threat models. 
In our experiments, LiRA illustrates the performance of LiRA under the threat model $\attack (\architecture, \eta)$, where the attacker knows the target model's architecture ($\architecture_\mathcal{T}$) and the associated hyperparameters ($\eta_\mathcal{T}$). Henceforth, it will serve as the baseline for evaluating the performance of other attacks proposed in this paper, noting that it is a more powerful attack and uses information not available to the other attacks.

\subsection{Testing Attacks in $\attack (\architecture)$ Setting}
One of the threat models discussed in the paper is $\attack (\architecture)$ in which the attacker's knowledge is restricted to the target model's architecture ($\architecture$). As mentioned in \Cref{tab:settings_and_attacks}, the attacks feasible in this setting include ACC-LiRA and KL-LiRA.

\Cref{fig:roc_plots_mia} presents the ROC curves depicting the success rate of our attack at different FPRs for each of the experiments. A summary of these plots is provided in
\Cref{fig:summary_plot_1} which demonstrates the efficacy of these attacks relative to LiRA over all the experiments at different $\fpr$s. \Cref{fig:roc_plots_mia} shows that KL-LiRA consistently performs on par with LiRA across all settings even though the attacker is provided with less information about the training process. On the other hand, ACC-LiRA performs poorly when compared to KL-LiRA and LiRA, especially at very low $\fpr$s. At $\fpr = 0.1\%$, KL-LiRA's average $\tpr$ is near equal to that of LiRA showing that training the shadow models to emulate the output loss distribution of the target model (as done in KL-LiRA) leads to no loss of efficiency for the attack. On the other hand, the average $\tpr$ for ACC-LiRA drops to $25\%$ of the corresponding $\tpr$ of LiRA. This shows that prioritizing the utility of shadow models yields a subpar attack.

\begin{figure}[htb]
    \centering
    \includegraphics[width=0.35\textwidth]{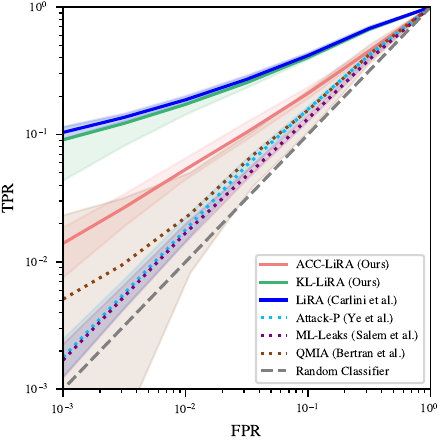}
    \caption{Comparing MIA efficiency (TPR) for KL-LiRA and ACC-LiRA against MIAs that do not rely on training shadow models, ML-Leaks \cite{DBLP:conf/ndss/Salem0HBF019}, Attack-P \cite{ye_enhanced_2021}, and QMIA \cite{DBLP:conf/nips/BertranT0KM023}. For different versions of LiRA, we use $\numbershadowmodels = 128$ per target model. This allows LiRA and its variants to have a distinguishable advantage over attacks not using any shadow models. Each target model is trained on CIFAR10 ($S=50$) with ViT-B + Head architecture.}
    \label{fig:diff_mias_comparison}
\end{figure}

 \subsection{Comparison with Shadow-Model-Free MIAs}
\label{appendix:no_shadow_model_mias}
In the $\attack (\architecture)$ setting, where the attacker has no knowledge of the target model's hyperparameters, MIAs that do not require training multiple shadow models (shadow-model-free MIAs) such as ML-Leaks \cite{DBLP:conf/ndss/Salem0HBF019}, Attack-P \cite{ye_enhanced_2021} or QMIA \cite{DBLP:conf/nips/BertranT0KM023} could be used to circumvent the need to find optimal hyperparameters for training the shadow models. In \Cref{fig:diff_mias_comparison}, we compare our proposed attacks against such shadow-model-free MIAs as baselines. We find that both KL-LiRA and ACC-LiRA outperform the shadow-model-free baselines in terms of the their performance at low $\fpr$s. This is in line with the observations made by Zarifzadeh et al. \cite{zarifzadeh_low-cost_2023} where shadow-model-free MIAs were significantly worse than the shadow-model-based MIAs in the low $\fpr$ regime.

\begin{figure}
   \centering
    \subfloat[Target: R-50, Shadow: ViT-B]{
    \centering 
    \includegraphics[width=0.46\columnwidth]{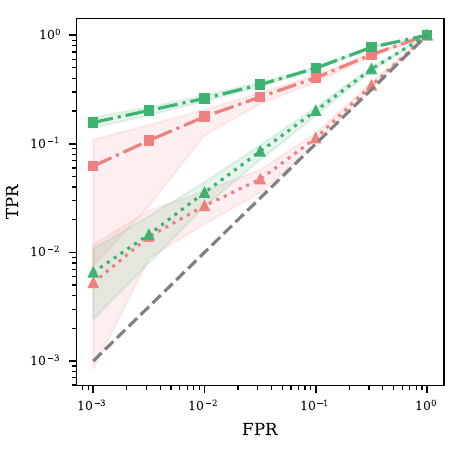}}
    ~
    \subfloat[Target: ViT-B, Shadow: R-50]{
    \includegraphics[width=0.46\columnwidth]{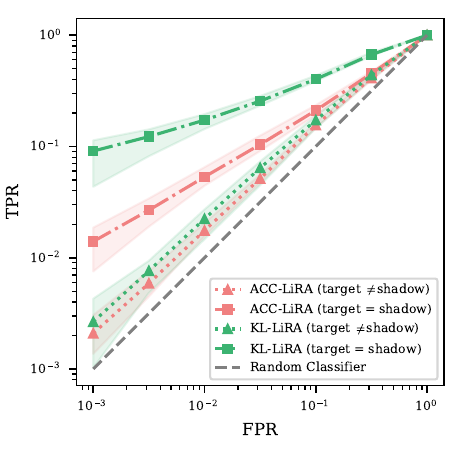}}

    \caption{Performance for KL-LiRA and ACC-LiRA in the $\bb$ setting against non-private models. For these plots, we assume the attacker has no knowledge of the target model's architecture and thus trains the shadow models using a different architecture. We used Head parameterization for these experiments. For comparison, we also depict the MIA vulnerability for the attacks when the target and shadow architecture are the same. All the attacks were run using $\numbershadowmodels=128$ shadow models. The mismatch between target and shadow architectures adversely effects the performance of the attacks showing that they are sensitive to the choice of shadow architecture.}
    \label{fig:diff_arch_plot}
\end{figure}

\subsection{Testing Attacks in $\bb$ Setting}
$\bb$  represents the most restrictive threat model. The attacker only has access to the final model with no additional information about the training process.
\Cref{fig:diff_arch_plot} demonstrates that the efficiency of ACC-LiRA and KL-LiRA decreases in the $\bb$ threat model.  Both attacks suffer due to the mismatch between the target and shadow model's architectures. The effect is particularly significant at low $\fpr$s. Their efficacy at $\fpr=0.1\%$ degrades by nearly $93\%$ of their performance when target and shadow models are trained with the same architecture.
Carlini et al.~\cite{carlini_membership_2021} also show that LiRA perform best when the shadow models' are trained with the same architecture as the target model. Changing the shadow model's architecture is expected to alter the performance of the attacks in the $\bb$ setting.
Nevertheless, KL-LiRA retains its advantage over ACC-LiRA as a more successful attack. The average $\tpr$ at $\fpr=0.1\%$ for KL-LiRA in this setting is $75\%$ higher than ACC-LiRA's.

\subsection{Testing Attacks on Differentially Private Models}

\begin{figure}
    \centering     
    \includegraphics[width=0.35\textwidth]{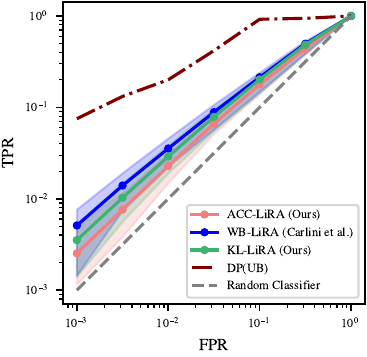}
    \caption{The plot above summarizes the performance of the attacks against models trained with $(\varepsilon = 8, \delta=10^{-5})$-DP as calculated over all the experiments and runs. DP(UB) represents the theoretical upper bound on the $\tpr$ at $\varepsilon = 8$. For all the experiments, we run the attacks using $\numbershadowmodels=128$ shadow models except for FiLM which uses $\numbershadowmodels=64$ shadow models. Using DP offers an effective defence against different versions of LiRA.
    A detailed version of this plot is available at \Cref{fig:roc_plots_mia_dp} which depicts the ROC plot for each experiment.}
    \label{fig:summary_plot_dp_8}
\end{figure}

\begin{figure}[htb]
    \centering
    \includegraphics[width=0.5\textwidth]{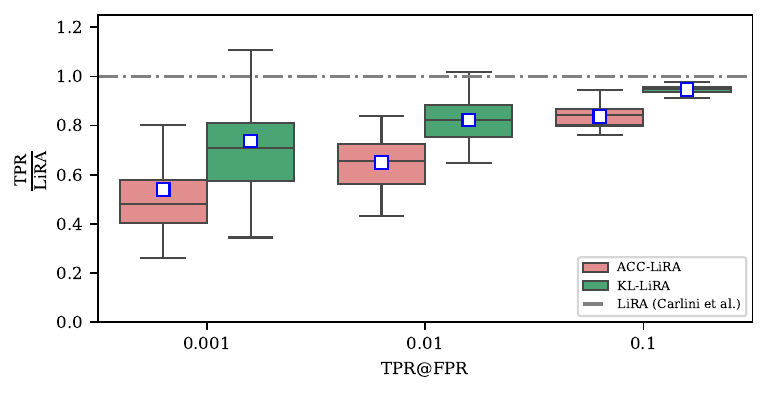}
    \caption{Overview of relative attacker performances against models trained with $(\varepsilon=8,\delta=10^{-5})$-DP. We normalize the MIA vulnerability ($\tpr$) at fixed $\fpr$ for KL-LiRA and ACC-LiRA by the corresponding $\tpr$ for LiRA for each experiment and run. It can be seen that KL-LiRA is a better choice for attack than ACC-LiRA under DP.}
    \label{fig:norm_summary_plot_dp_8}
\end{figure}

In their paper on LiRA, Carlini et al. \cite{carlini_membership_2021} demonstrated that DP can be used as an effective defense against LiRA. From \Cref{fig:summary_plot_dp_8}, it can be seen that this remains true for the other versions of LiRA that we use in this paper. Privacy leakage ($\tpr$) for none of the attacks comes close to the theoretical upper bound for $\tpr$ under DP (depicted by DP(UB) in the plot). 
We compute the upper bound using the method of~\cite{Kairouz2015Composition} described in  \Cref{thm:Kairouz_bound}. Using a fixed $\delta = 10^{-5}$ and the associated $\varepsilon$ will not yield an optimal bound. To improve this, we  compute the tightest bound over all $\varepsilon(\delta)$ values satisfied by the algorithm, as evaluated by the privacy accountant. \Cref{fig:summary_plot_dp_8} tells us that it suffices to use a relatively high privacy budget of $\varepsilon=8$ to defend against KL-LiRA or ACC-LiRA.

\Cref{fig:norm_summary_plot_dp_8} depicts the relative success of ACC-LiRA and KL-LiRA compared to LiRA when the models are trained with ($\varepsilon=8,\delta=10^{-5}$)-DP. Against models trained with DP, KL-LiRA was found to be a superior attack compared to ACC-LiRA even though its performance suffers relative to LiRA from the non-DP setting. At $\fpr=0.1\%$, average $\tpr$ for KL-LiRA is 1.4$\times$ higher than ACC-LiRA's average $\tpr$ for $\varepsilon = 8$.

\begin{figure}[htb]
    \centering
    \includegraphics[width=0.75\columnwidth]{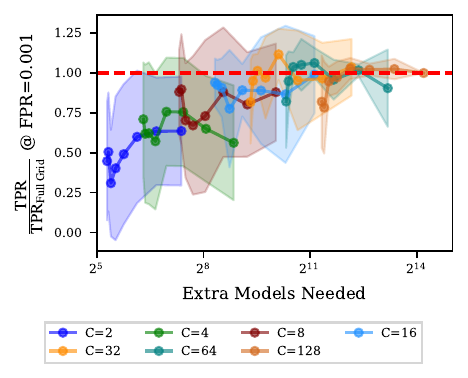}
    \caption{The plot depicts the $\tpr$ for KL-LiRA as we vary the number of candidate hyperparameters ($C$) and the shadow models trained for using each candidate ($N$) relative to the $\tpr$ of the Full Grid ($C=N=M$) at $\fpr=0.001$. Along the x-axis, we depict the extra models needed for hyperparameter selection in KL-LiRA, $(C\times T) + C\times(N-1) - N$, where $T$ is the number of HPO trials per candidate ($T=20$). The error bars depict the standard deviation around the average $\tpr$. To find the optimal hyperparameters that yield performance comparable to the Full Grid, we will need $C \ge 16, N \ge 1$. We ran the attack against non-a private target model trained on CIFAR100 ($S=100$) with ViT-B + Head architecture.}
    \label{fig:optimizing_kl_lira}
\end{figure}

\subsection{Optimizing KL-LiRA}

To attack a single target model using KL-LiRA, the attacker needs to find a set of $C$ candidate hyperparameters, $\{\eta_j\}_{j=1}^{C}$. This set should be large enough to allow the attacker to find the hyperparameters that best approximate the behaviour of target model's hyperparameters. We also need to optimize $N$, the number of shadow models that the attacker will need to train for each of the candidate hyperparameters which are used for hyperparameter selection in \Cref{alg:kl-lira}.  Assuming that the attacker has access to $M$ shadow data sets, a maximally efficient attack would use all the $M$ data sets for the hyperparameter selection process, that is, $C=N=M$. We refer to this as the Full Grid KL-LiRA. However, as mentioned in \Cref{subsec:compcostattacks}, the cost for running KL-LiRA scales with $C$ and $N$. Here, we tried to estimate how many extra models (apart from the $\numbershadowmodels$ shadow models that would be trained for the attack) will be needed to find the optimal hyperparameters for KL-LiRA to attack one target model such that the attack's success rate is comparable to its success rate when using the Full Grid. For these experiments, we use $T=20$ HPO trials. \Cref{fig:optimizing_kl_lira} shows that for matching the performance of the Full Grid ($C=N=M$), the attackers needs to have $C \ge 16$ candidate hyperparameters and $N \ge 1$ models trained using each of the candidates.

\subsection{Additional Experiments}
We examine the effect of choosing a weaker pretrained architecture, namely, WideResNet-50 (WR-50) in \Cref{appendix:weaker_pretrained_backbone}. We find that both KL-LiRA and ACC-LiRA are relatively close in performance to LiRA in this setting, indicating this to be an easy setting for attacks.
We also analyze the performance of KL-LiRA against models trained from scratch in \Cref{appendix:train_from_scratch}. KL-LiRA outperforms ACC-LiRA in this setting and is closest to the performance of LiRA.

\section{Empirical Privacy Leakage Due To HPO: Methods}
\label{sec:hpo_leakage}

\begin{figure*}[htb]
    \centering
    \includegraphics[width=0.85\textwidth]{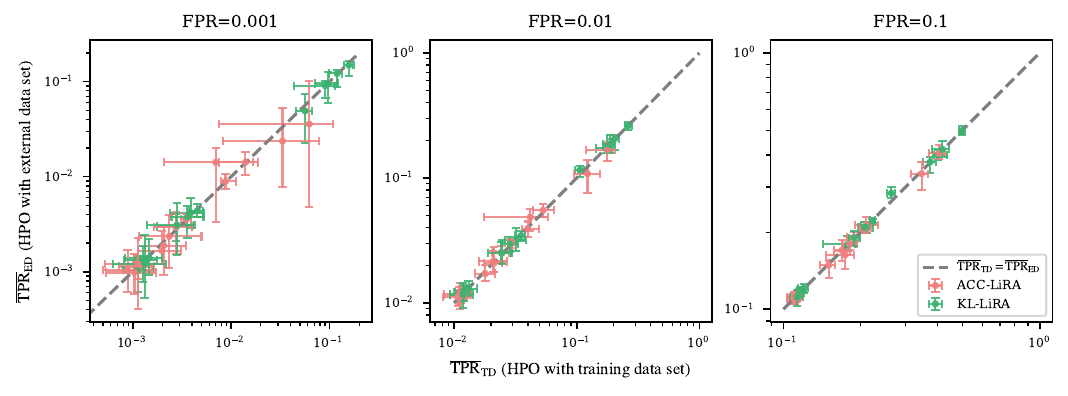}
    \caption{The plot above compares average MIA vulnerability ($\overline{\tpr}$) when the HPO is done using the training data (TD-HPO) against the vulnerability of models for which the HPO is done using external data set (ED-HPO) for different experiments and privacy levels ($\varepsilon = \infty,8, 1$) at fixed $\fpr$. The error bars denote the Clopper--Pearson confidence intervals for the $\tpr$ along the x and y-axis. If there is no significant additional leakage observed for TD-HPO, the points are expected to fall on the diagonal representing equal privacy leakage for TD- and ED-HPO. If TD-HPO is a more vulnerable setting, the points are expected to fall on the right of the diagonal. Most of the points align with the diagonal showing no significant difference between MIA vulnerability for the two settings.}
    \label{fig:summary_plot_td_vs_ed}
\end{figure*}

Next, we study the impact of HPO under DP on the MIA vulnerability of the final released classification model under the assumption that the hyperparameters are not released.

We evaluated the privacy leakage of HPO by comparing the MIA vulnerability of two approaches: a training-data-based HPO (TD-HPO) and an HPO based on an external data set disjoint from the training data (ED-HPO). In
TD-HPO, we use the training data to find the optimal hyperparameters and perform the final training of the model under DP. In ED-HPO, we find the hyperparameters using a data set disjoint from the training data
and use the training data only for the final DP training. Given that ED-HPO does not use the training data for HPO, the hyperparameters come without a privacy cost (i.e. $(0,0)$-DP) relative to the training data. Therefore, the ED-HPO setting formally satisfies the provided $(\varepsilon, \delta)$-DP guarantee, while the best known guarantee for TD-HPO would be significantly weaker. Now, to test whether TD-HPO incurs a higher privacy cost, we compare its MIA vulnerability to that of ED-HPO.

In our experiments, we use the same MIA-Grid of shadow models for both the settings, with the only difference being that for ED-HPO, the target model's HPO is done using an external data set. For ED-HPO, we subsample $M+1$ data sets, $\{\data^e_i\}_{i=0}^{M}$, from $\mathbb{D}$ that are disjoint from the training data sets, to calculate an alternate set of hyperparameters for training the models in the ED setting, $\{\eta^e_i\}_{i=0}^{M}$. In ED-HPO, the training of the target models adheres to the following protocol,

\begin{equation}
\begin{split}
    \eta^e_i =
    \texttt{HPO}(\data^e_i), \\
    \model^e_{\data_i,\eta^e_i} = \texttt{TRAIN}(\architecture, \data_i, \eta^e_i), \\
    \text{ where } \data^e_i \cap \data_j = \emptyset \quad \forall i,j.
\end{split}
\end{equation}

As indicated previously, our primary concern lies in assessing the privacy risk associated with HPO under the assumption that the hyperparameters of the target model remain undisclosed. Accordingly, the experiments conducted in this section evaluate the MIA vulnerability through the use of KL-LiRA and ACC-LiRA, wherein the attacker can run the attack without prior knowledge of the target model's hyperparameters.

\begin{figure*}[htb]
    \centering    \includegraphics[width=0.9\linewidth]{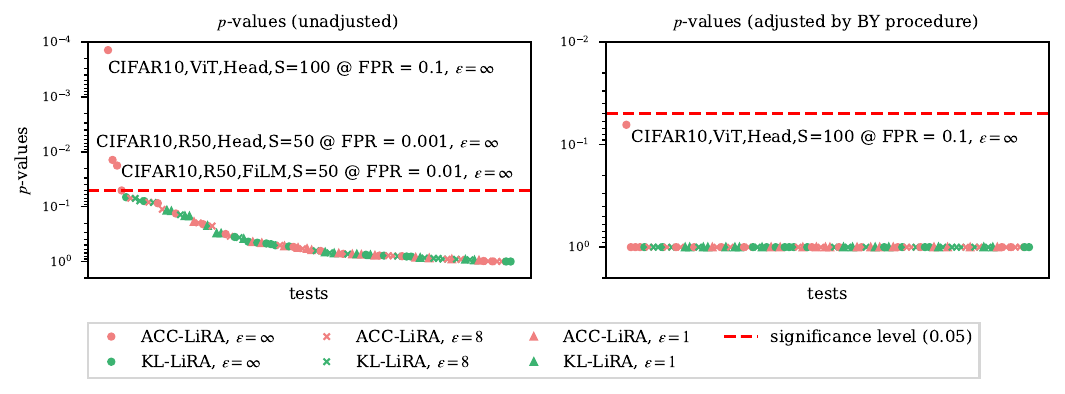}
    \caption{The plot above shows $p$-values (unadjusted and adjusted by Benjamini-Yekutieli (BY) procedure) on a negative logarithmic scale obtained for the t-tests comparing the distribution of $\tpr$ at fixed $\fpr$ $(0.001,0.01,0.1)$ for TD-HPO (where HPO is done using the training data), $\prob_{\text{TD}}(\tpr|\fpr)$, and ED-HPO (where HPO is done using external data), $\prob_{\text{ED}}(\tpr|\fpr)$, for all the experiments sorted in increasing order. The negative-logarithmic scale highlights the $p$-values that are significant as being above the horizontal dashed line. Additionally,
    we provide the details of the experiments for which the $p$-value is above the significance level or stands out among the adjusted $p$-values.}
    \label{fig:mahatten_plots_ttest_v2}
\end{figure*}

\section{Empirical Privacy Leakage Due To HPO: Results}
\label{sec:hpo_leakage_results}
\Cref{fig:summary_plot_td_vs_ed} summarizes the average MIA vulnerability for both TD-HPO and ED-HPO for different attacks at varying levels of privacy ($\varepsilon=\infty, 8, 1$) with $\delta=10^{-5}$. The vulnerability to MIA is very similar between TD and ED at all $\fpr$ levels, as observed by most of the points in \Cref{fig:summary_plot_td_vs_ed} being on or near the diagonal representing $\overline{\tpr}_{\text{TD}}=\overline{\tpr}_{\text{ED}}$. For the points farther from the diagonal, the error bars (representing the Clopper--Pearson confidence interval) are large and overlap with the diagonal. 
In \Cref{fig:summary_plot_td_vs_ed}, We provide a granular version of these plots for each $\varepsilon$ in \Cref{fig:td_vs_ed_by_dp} and \Cref{fig:td_vs_ed_by_dp_logscale}.

We also used paired statistical tests (t-test \cite{87464e34-37bc-3288-807c-e421e5a0d7a6} and permutation test \cite{c6bd8513-eac7-3eed-b41d-0c7ba1981dc2}) to compare the distribution of $\tpr$ at a given $\fpr$ for TD-HPO and ED-HPO over a range of experiments. We use the paired version of the tests because our data are in the form of a matched pair.

For each experiment, we aim to measure the additional vulnerability due to HPO using the training data (TD-HPO) over the vulnerability of the ED-HPO where the training data is not used for HPO for a given target model. Thus, our subject in each experiment (the target model) changes only in terms of how the HPO is performed. Paired tests are, in general, more powerful that the unpaired tests as they control for variability between samples by using the same participants or items, isolating the effect of the variable being tested.

For the tests, our null hypothesis ($H_0$) is that TD-HPO does not increase the MIA vulnerability of the training data compared to ED-HPO. All our tests are 1-sided, with the alternative hypothesis ($H_1$) being that TD-HPO leads to a higher vulnerability (observed as an increase in $\tpr$ at a given $\fpr$) than ED-HPO. We maintain the level of statistical significance for all tests $\alpha = 0.05$. For each of these tests, we record the observed $p$-values (\Cref{tab:ttest_table} and \ref{tab:permtest_table}).

Our statistical analysis involves multiple statistical tests, which increases the probability of encountering false discoveries (incorrect rejections of $H_0$). To account for these false discoveries, we employ the Benjamini--Yekutieli (BY) procedure~\cite{benjaminiyekutieli} which involves adjusting the $p$-values of individual tests to control the false discovery rate (FDR) at a given significance level $\alpha$. 
This method makes no assumptions about the correlations between different comparisons, which makes it suitable for our tests that can be correlated due to repeated use of the same data. However, the trade-off is reduced power, meaning it identifies fewer comparisons as discoveries. In other words, the method is conservative.

\Cref{fig:mahatten_plots_ttest_v2} shows the $p$-values of different tests and the corresponding BY-adjusted p-values on a logarithmic scale sorted in ascending order. 
We can see that the number of rejections of $H_0$ after controlling the FDR drops from three to zero for the paired t-test. This suggests that among our experiments, we do not see any examples of TD-HPO resulting in a statistically significant increase of MIA vulnerability compared to the ED-HPO. We observe exactly the same using permutation test: without FDR-control we reject the null three times and after adjusting we retain the null for all the tests (see \Cref{fig:mahatten_plots_permtest_v2}).
Interestingly, no DP experiment yields significant $p$-values even without the adjustment, as all the most significant cases are with non-DP models.

\section{Discussion}

\subsection{Importance of Choosing Fitting Hyperparameters for LiRA:}
 Using good hyperparameters to train the shadow models is critical to the performance of LiRA.  As observed in \Cref{sec:attack_results}, ACC-LiRA performs considerably worse than LiRA and KL-LiRA, especially at low $\fpr$. 
 It can be seen from \cref{fig:loss_dist_plot} that the variance of the IN and OUT distributions for data samples is much larger for ACC-LiRA as compared to LiRA or KL-LiRA.  ACC-LiRA differs from the other two attacks in its approach to choose optimal hyperparameters to train the shadow models. We suspect that the large variance of the distributions is due to each shadow model being trained with a different set of hyperparameters with the aim to maximize their utility on the corresponding shadow data sets. 
 
Additionally, \Cref{fig:relation_hps_mias} shows that the hyperparameters (learning rate and batch size) found using ACC-LiRA are not close to the target hyperparameters. Whereas for KL-LiRA, the learning rate used for training the shadow models is similar to the target learning rate. This makes it difficult for ACC-LiRA to distinguish between the 2 distributions thus inhibiting its ability to label the data samples as IN or OUT with confidence.
 
\subsection{Accounting for Privacy Leakage due to HPO Under DP:} The evaluation of privacy of machine learning models is composed of two components: formal DP bounds that give an upper bound and practical attacks that give a lower bound on the privacy loss.
DP bounds are often based on simplifications that risk making them loose, while with attacks it is difficult to rule out the existence of an even stronger attack.
It has been shown that DP bounds for DP-SGD are tight under certain conditions when releasing the whole sequence of iterates \cite{DBLP:journals/corr/abs-2101-04535}.
There are very few tools for computing tight DP bounds when not all internal steps are released.
\cite{Ye2022HiddenState} and \cite{Altschuler2022MoreIters} show that in well-behaved problems the cumulative privacy loss of an iteration can be bounded, but these methods provide no help in bounding the extra privacy loss in more complex settings such as due to HPO.
None of our statistical tests for DP experiments yield a $p$-value (with or without adjustment) that would lead to a rejection of the null hypothesis (that the TD-HPO and ED-HPO result on average in equal MIA vulnerability). Our results suggest that the privacy bounds provided by current DP HPO methods are likely to be loose.

It is worth noting, that while our statistical tests fail to reject the null hypothesis, this absence of evidence does not imply evidence of absence. However, as we saw in \cref{fig:summary_plot_td_vs_ed}, the paired vulnerabilities across different settings are all fairly close to the diagonal. Hence while there might be some difference in the vulnerability distributions between the TD and ED, the effect seems to be rather limited.

\textbf{Limitations} From \Cref{fig:optimizing_kl_lira} it can be seen that KL-LiRA’s success relies on training multiple shadow models for the task. Training shadow models can become computationally intensive when attempting to target larger models. For example, if running LiRA requires training $256$ shadow models, for KL-LiRA, the attacker will need to train at least $\sim 1000$ shadow models for an efficient attack.

In our experiments, the shadow models’ training data set partially overlaps with the target model’s training data set. This assumption strengthens the attacks discussed in our paper as it corresponds to a stricter threat model. It remains to be seen how the performance of KL-LiRA would be impacted if the training data of shadow models and the target model are disjoint or belong to different distributions. This is a limitation which we inherit from LiRA.

Furthermore, in our experiments we restrict the tunable hyperparameters to the training batch size and learning rate as well as possible additional DP-SGD parameters. It is yet unclear how the effectiveness of the proposed attacks might change as the number of tunable hyperparameters increases. We briefly explore this in \Cref{appendix:diff_set_hypers} where we expand the set of unknown hyperparameters to include the number of training epochs.

\section{Conclusion}
In this paper, we demonstrated that score-based MIAs that use shadow models (e.g. LiRA) do not require prior knowledge of target model's hyperparameters as a necessary pre-condition for success, provided that an alternate set of hyperparameters is chosen to train the shadow models to optimize the efficiency of the attack. 
To facilitate this, we introduced KL-LiRA which selects hyperparameters for shadow model training based on similarity between the output distributions of the target model and shadow models. In our experiments, KL-LiRA's performance was very similar to an attack using target model's hyperparameters against non-DP models, thereby showing that keeping the hyperparameters secret provides no additional privacy protection for a non-DP model. For models trained with DP, KL-LiRA performed slightly worse than an attack using target model's hyperparameters. We also studied whether using training data for HPO under DP leads to privacy leakage that is significant enough to be detected by an empirical privacy attack such as MIA. We found no evidence that suggests that using training data for HPO would lead to significant additional privacy leakage in the transfer learning setting. 

Source code for our experiments is available at: \url{https://github.com/DPBayes/HPs_In_Score_Based_MIAs}. 
We implement the attacks by adapting the code provided by prior works' authors \footnote{\url{https://github.com/cambridge-mlg/dp-few-shot } (Tobaben et al. \cite{tobaben_efficacy_2023})}\footnote{\url{https://github.com/tensorflow/privacy/tree/master/research/mi_lira_2021}(Carlini et al. \cite{carlini_membership_2021})}.

\section*{Acknowledgment}

This work was supported by the Research Council of Finland (Flagship programme: Finnish Center for Artificial Intelligence, FCAI, Grant $356499$ and Grant $359111$), the Strategic Research Council at the Research Council of Finland (Grant $358247$) as well as the European Union (Project $101070617$). Views and opinions expressed are however those of the author(s) only and do not necessarily reflect those of the European Union or the European Commission. Neither the European Union nor the granting authority can be held responsible for them. This work has been performed using resources provided by the CSC– IT Center for Science, Finland (Project $2003275$).

\bibliographystyle{IEEEtran}
\bibliography{citations}

\appendix

\counterwithin*{figure}{part}
\stepcounter{part}
\renewcommand{\thefigure}{A\arabic{figure}}
\setcounter{figure}{0}
\counterwithin*{table}{part}
\stepcounter{part}
\renewcommand{\thetable}{A\arabic{table}}
\setcounter{table}{0}
\renewcommand{\thesubsection}{A.\arabic{subsection}} 
\subsection{Hyperparameter Optimization}
\label{appendix1}

For a given input data set, $\mathcal{\data}$, we perform hyperparameter optimization (HPO) using 70$\% $ of $\mathcal{\data}$ for training and the remaining 30$\%$ is used for validation. 
HPO is done over 20 iterations using the BoTorch estimator \cite{balandat_botorch_2019} with Optuna \cite{akiba_optuna_2019} to find a set of optimal hyperparameters for the optimizer that yield the highest accuracy on the validation split. In our experiments, we regard the noise multiplier as a hyperparameter to be computed using the target $(\varepsilon, \delta)$-DP privacy budget. 
Hyperparameters and their corresponding ranges used for HPO are detailed in \Cref{tab:hyperparam_ranges}.

\begin{table}[ht!]
    \renewcommand{\arraystretch}{1.1}
    \caption{Hyperparameter and their ranges used for the Bayesian Optimization with Optuna.}
    \label{tab:hyperparam_ranges}
    \centering
        \adjustbox{max width=\columnwidth}{\begin{tabular}{l|cc}
            \hline
            \multicolumn{1}{l}{\textbf{Hyperparameter}} & \multicolumn{1}{l}{\textbf{Lower Bound}} & \multicolumn{1}{l}{\textbf{Upper Bound}} \\
            \hline
            Batch Size & \multicolumn{1}{r}{10} & \multicolumn{1}{r}{$|\mathcal{D}|$} \\
            Gradient Clipping Norm & \multicolumn{1}{r}{0.2} & \multicolumn{1}{r}{10} \\
            Learning Rate & \multicolumn{1}{r}{1e-7} & \multicolumn{1}{r}{1e-2} \\
            Noise Multiplier & \multicolumn{2}{c}{Based on target $\epsilon$} \\
            Epochs & \multicolumn{2}{c}{Constant, $=40$} \\
            \hline   
        \end{tabular}}
\end{table}

\begin{figure}[htb]
    \centering
    \includegraphics[width=0.7\linewidth]{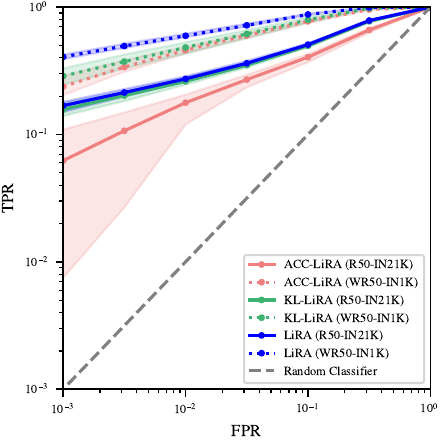}
    \caption{MIA efficiency ($\tpr$) against WR-50 + Head target models when compared to MIA efficiency against R-50 + Head target models trained on CIFAR-10 ($S=50$). WR-50 is trained on ImageNet-1K making it a weaker pretrained model than R-50 trained on ImageNet-21K. Performance of all versions of LiRA improve with a weaker pretrained backbone implying this to be an easy setting for attacks. For each of these attacks, we use $M=128$ shadow models per target model. Each target model is trained on CIFAR10 ($S=50$) data set.}
    \label{fig:wr50_plot}
\end{figure}

\begin{figure}[htb]
    \centering
    \includegraphics[width=0.75\linewidth]{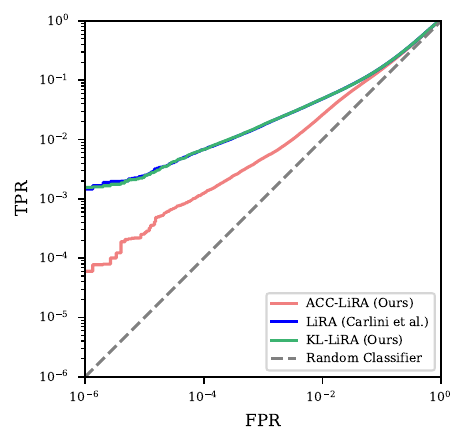}
    \caption{Evaluating attack performance against models trained from scratch using MNIST data set calculated over $100$ target models. Each target model was attacked using $99$ shadow models. Similar to deep transfer learning setting, we find that KL-LiRA is the optimal choice for attack when the attacker has no knowledge of the target model's hyperparameters.}
    \label{fig:from_scratch_plot}
\end{figure}

\subsection{Attacking Models Using Weaker Pretrained Backbone}
\label{appendix:weaker_pretrained_backbone}
To evaluate the effect of using a weaker pretrained backbone for the target model, we fine-tune Wide-ResNet-50 (WR-50) with $68.9$M parameters \cite{DBLP:conf/bmvc/ZagoruykoK16} pretrained on ImageNet-1K \cite{russakovsky_imagenet_2014} using Head as parameterization as our target models for 120 training epochs. In our experiment, we consider the $\attack(\architecture)$ setting. Therefore, the shadow models have the same architecture as the target models. The results are shown in \cref{fig:wr50_plot}. Using a pretrained model trained on a subset of ImageNet-21K tends to enhance the attack efficiency of all 3 variants of LiRA indicating that this is an easy setting for attacks. Evidently, the performance of KL-LiRA relative to LiRA suffers when a weaker pretrained model is used for transfer learning but not significantly. $\tpr$ for KL-LiRA is $0.85\times$ of the $\tpr$ for LiRA at $0.1\%$ $\fpr$ with WR-50 while it is $0.88\times$ of the $\tpr$ for LiRA at $0.1\%$ $\fpr$ with R-50 \cite{kolesnikov_big_2019}.

\subsection{Attacking Models Trained From Scratch}
\label{appendix:train_from_scratch}
To evaluate the efficiency of KL-LiRA beyond the deep transfer learning setting, we train a CNN with a convolutional layer (with 16 filters) followed by a fully connected layer (with 10 units) from scratch using MNIST data set \cite{DBLP:journals/pieee/LeCunBBH98}. We train 100 such models as the target models for the attack. For training from scratch we run 100 HPO trials using Optuna for each target model. Our preliminary analysis as shown in \Cref{fig:from_scratch_plot} demonstrates that KL-LiRA manages to match the attack success rate ($\tpr$) of LiRA at low $\fpr$s and is a superior choice of attack to ACC-LiRA in this setting. 

\begin{figure}[htbp]
    \centering
    \includegraphics[width=0.7\linewidth]{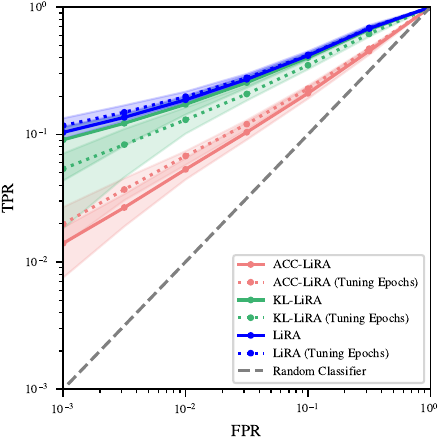}
    \caption{Impact of adding epochs to the set of hyperparameters unknown to the attack in $\attack(\architecture)$ threat model on the MIA efficiency. Increasing the set of unknown hyperparameters decreases the efficiency of KL-LiRA with respect to LiRA. For comparison, we also depict the performance of different attacks with fixed epochs ($= 40$) as baseline. In the latter setting, the optimizer hyperparameters unknown to the attacker include learning rate and batch size. For each of these attacks, we use $M=128$ shadow models per target model. Each target model is trained on CIFAR10 ($S=50$) with ViT-B + Head architecture.}
    \label{fig:var_epochs}
\end{figure} 

\subsection{Varying the Number of Unknown Hyperparameters in HPO}
\label{appendix:diff_set_hypers}
In this section, we briefly explores how KL-LiRA fairs when we expand the set of hyperparameters unknown to the attacker to include the exact number of epochs used for training the target model. During HPO, we maintain a range of $\{1,200\}$ for the number of training epochs.
From \Cref{fig:var_epochs}, it can be seen that KL-LiRA continues to be the better approach to find optimal hyperparameters for training the shadow models as the pool of hyperparameters not known to the attack expands to include the number of epochs used to train the target model. The power of KL-LiRA suffers with a larger set of unknown hyperparameters which requires further investigation in the future.
$\tpr$ for KL-LiRA (when tuning for training epochs) is $0.62\times$ lower than $\tpr$ for KL-LiRA (without tuning for training epochs) at $0.1\%$ $\fpr$.

\newpage
\subsection{Additional Figures and Tables}

\begin{figure*}[htb]
    \centering
    \subfloat[CIFAR10, ViT-B + Head, $S = 100$]{

    \includegraphics[width=0.3\textwidth]{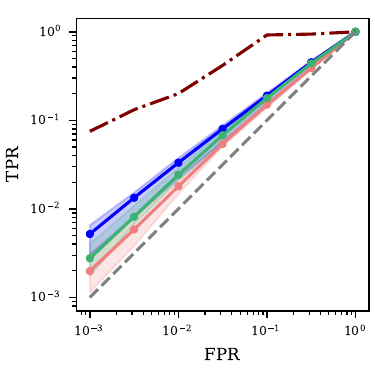}}
    ~
    \subfloat[CIFAR10, ViT-B + Head, $S = 50$]{

    \includegraphics[width=0.3\textwidth]{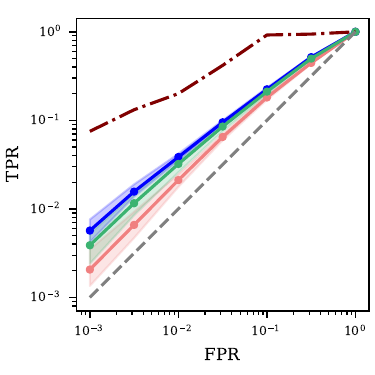}}
    \subfloat[CIFAR100, ViT-B + Head, $S=100$]{

    \includegraphics[width=0.3\textwidth]{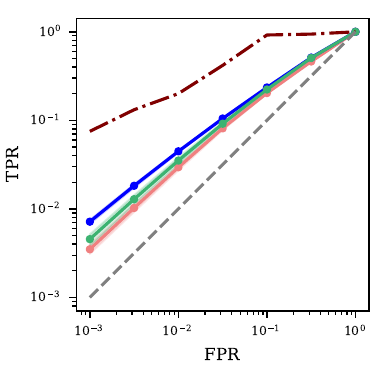}}
    \\
    \subfloat[CIFAR10, R-50 + Head, $S = 50$]{

    \includegraphics[width=0.3\textwidth]{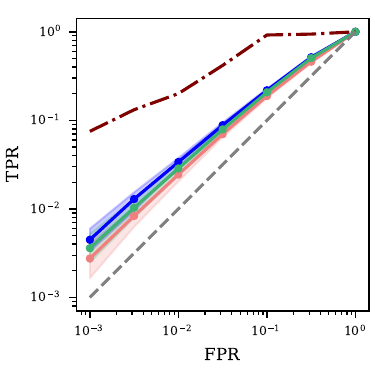}}
    \subfloat[CIFAR10, R-50 + FiLM, $S=50$]{

    \includegraphics[width=0.3\textwidth]{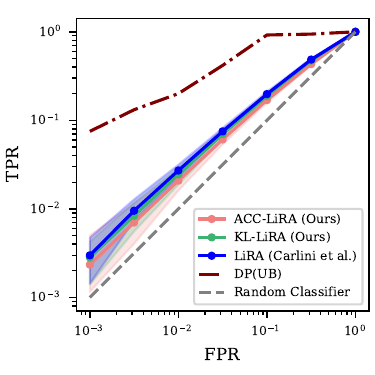}}
    \caption{Attack performances against models protected by DP ($\varepsilon=8, \delta=10^{-5}$). ROC curves depict the performance of different attacks for individual experiments computed over 10 runs. All the plots are generated using 128 shadow models except for FiLM which uses 64 shadow models. The error bars represent the Clopper--Pearson confidence interval associated with the estimated $\tpr$ at fixed $\fpr$. Using DP has adverse effect on the attacks' success rate.  Under DP, LiRA is the most powerful attack followed by KL-LiRA and ACC-LiRA.}
    \label{fig:roc_plots_mia_dp}
\end{figure*}
\begin{figure*}[htb]
    \centering
    \subfloat[CIFAR10, ViT-B + Head, $S = 100$]{

    \includegraphics[width=0.3\textwidth]{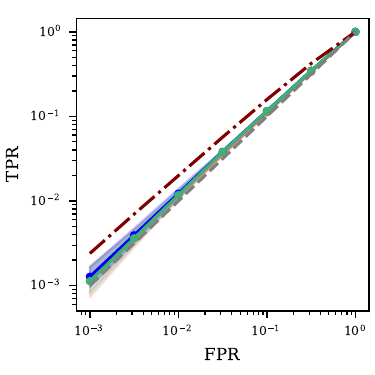}}
    ~
    \subfloat[CIFAR10, ViT-B + Head, $S = 50$]{

    \includegraphics[width=0.3\textwidth]{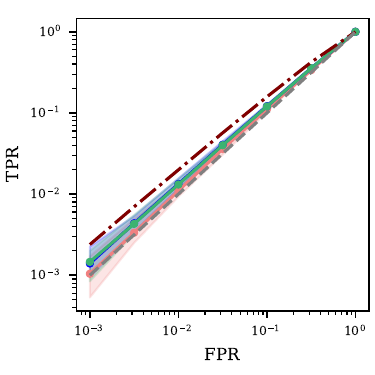}}
    \subfloat[CIFAR100, ViT-B + Head, $S=100$]{

    \includegraphics[width=0.3\textwidth]{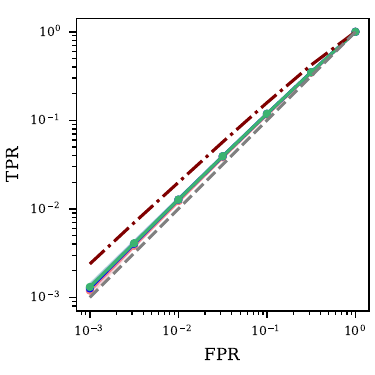}}
    \\
    \subfloat[CIFAR10, R-50 + Head, $S = 50$]{

    \includegraphics[width=0.3\textwidth]{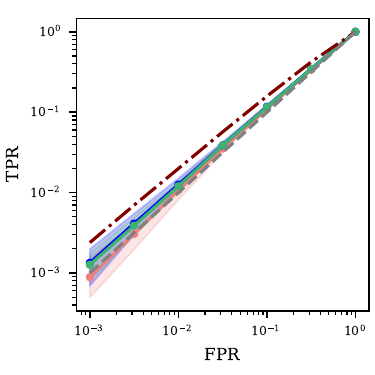}}
    \subfloat[CIFAR10, R-50 + FiLM, $S=50$]{

    \includegraphics[width=0.3\textwidth]{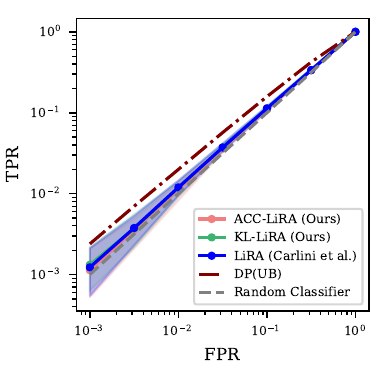}}
    \caption{Attack performances against models protected by DP ($\varepsilon=1, \delta=10^{-5}$). ROC curves depict the performance of different attacks for individual experiments computed over 10 runs. All the plots are generated using 128 shadow models except for FiLM which uses 64 shadow models. The error bars represent the Clopper--Pearson confidence interval associated with the estimated $\tpr$ at fixed $\fpr$. A moderate privacy level of $\varepsilon=1$ reduces the power of the attacks close to random guessing.}
    \label{fig:roc_plots_mia_dp_1}
\end{figure*}

\begin{figure*}
    \centering
    \subfloat[With $\varepsilon = \infty$]{
    \includegraphics[width=\linewidth]{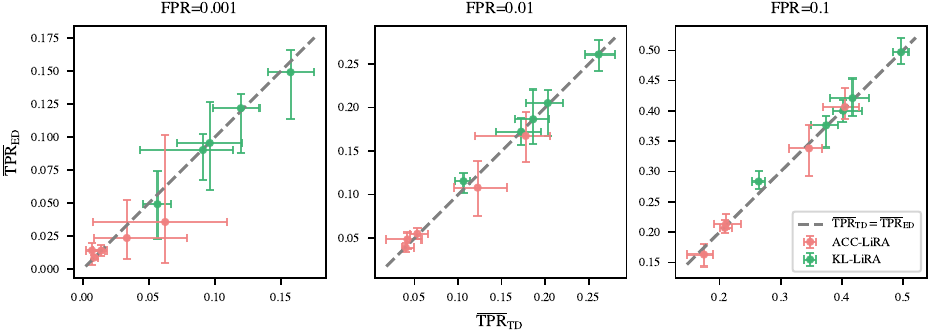}}
    \\
    \subfloat[With $\varepsilon = 8$]{
    \includegraphics[width=\linewidth]{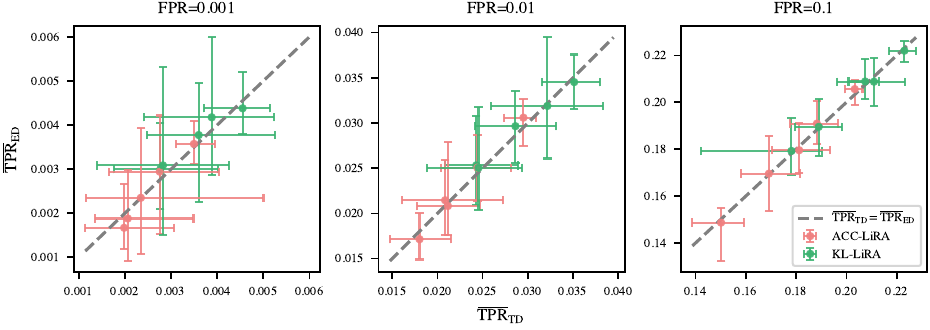}}
    \\
    \subfloat[With $\varepsilon = 1$]{
    \includegraphics[width=\linewidth]{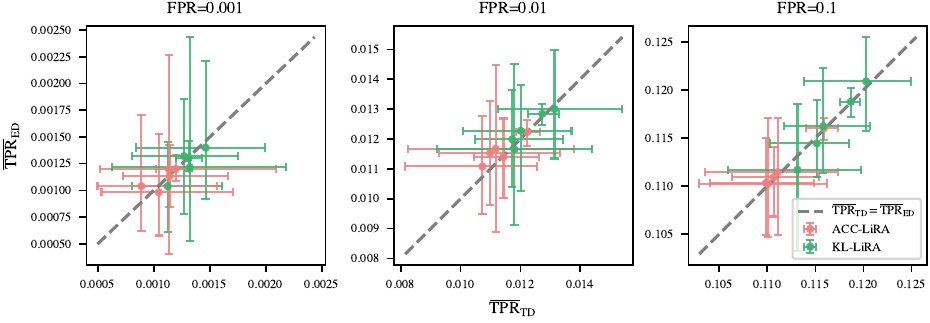}}
    \caption{The plot compares average MIA vulnerability ($\overline{\tpr}$) when the HPO is done using the training data (TD-HPO) against the vulnerability of models for which the HPO is done using external data set (ED-HPO) for different experiments grouped by privacy levels ($\varepsilon = \infty,8, 1$) at fixed $\fpr$. The error bars denote the Clopper--Pearson confidence intervals for the $\tpr$ along the x and y-axis. If there is no significant leakage observed for TD-HPO, the points are expected to fall on the diagonal representing equal privacy leakage for TD and ED-HPO. If TD-HPO is a more vulnerable setting, the points are expected to fall on the right of the diagonal. Most of the points align with the diagonal showing no significant difference between MIA vulnerability for the two settings.}
    \label{fig:td_vs_ed_by_dp}
\end{figure*}

\begin{figure*}
    \centering
    \subfloat[With $\varepsilon = \infty$]{
    \includegraphics[width=\linewidth]{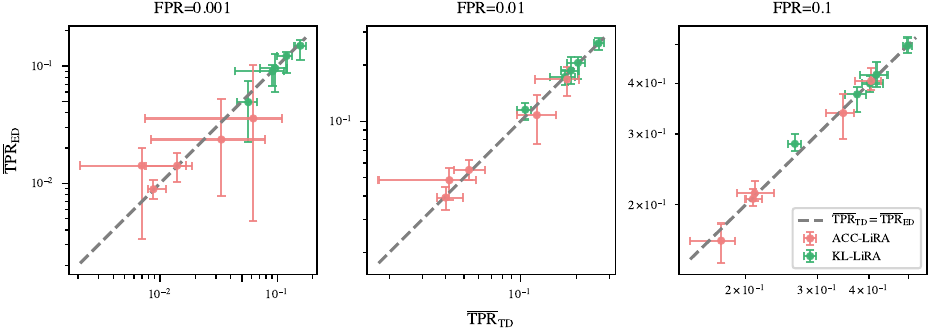}}
    \\
    \subfloat[With $\varepsilon = 8$]{
    \includegraphics[width=\linewidth]{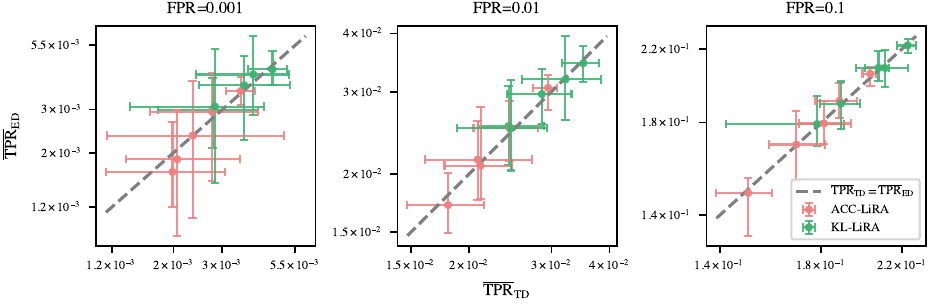}}
    \\
    \subfloat[With $\varepsilon = 1$]{
    \includegraphics[width=\linewidth]{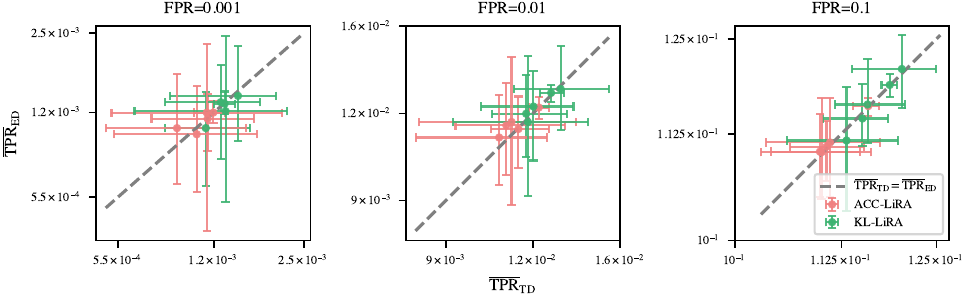}}
    \caption{The plot compares average MIA vulnerability ($\overline{\tpr}$) when the HPO is done using the training data (TD-HPO) against the vulnerability of models for which the HPO is done using external data set (ED-HPO) for different experiments and privacy levels ($\varepsilon = \infty,8, 1$) at fixed $\fpr$ on a logarithmic scale. The error bars denote the Clopper--Pearson confidence intervals for the $\tpr$ along the x and y-axis. If there is no significant leakage observed for TD-HPO, the points are expected to fall on the diagonal representing equal privacy leakage for TD and ED-HPO. If TD-HPO is a more vulnerable setting, the points are expected to fall on the right of the diagonal. Most of the points align with the diagonal showing no significant difference between MIA vulnerability for the two settings.}
    \label{fig:td_vs_ed_by_dp_logscale}
\end{figure*}

\begin{table*}[!htb]
\centering
\caption{Paired 1-sided student's t-test (with $H_1: \overline{\tpr}_{\text{TD}} > \overline{\tpr}_{\text{ED}}$) table comparing $\mathbb{P}_{\text{TD}}(\tpr|\fpr)$ and $\mathbb{P}_{\text{ED}}(\tpr|\fpr)$. $\overline{\tpr}_{\text{TD}}$ denotes the average $\tpr$ when HPO for the target model is done using the training data (TD-HPO) and $\overline{\tpr}_{\text{ED}}$ denotes the average $\tpr$ when HPO is done using an external dat set (ED-HPO).   $\overline{\Delta\tpr}$ represents $(\overline{\tpr_{\text{TD}} - \tpr_{\text{ED}}}) \times 10^{-4}$.} 
\label{tab:ttest_table}
\begin{center}
\adjustbox{max width=\textwidth}{\begin{tabular}{lllrclccccccccc}
\toprule
\multicolumn{1}{c}{\multirow{2}{*}{\textbf{Dataset}}} & \multicolumn{1}{c}{\multirow{2}{*}{\textbf{Model}}} & \multicolumn{1}{c}{\multirow{2}{*}{\textbf{Config}}} & \multicolumn{1}{c}{\multirow{2}{*}{\textbf{\boldmath $S$}}} &  \multicolumn{1}{c}{\textbf{\boldmath $\varepsilon$}}  & \multicolumn{1}{c}{\multirow{2}{*}{\textbf{MIA}}} & \multicolumn{3}{c}{\textbf{\boldmath $\fpr=10^{-3}$}} & \multicolumn{3}{c}{\textbf{\boldmath $\fpr=10^{-2}$}} & \multicolumn{3}{c}{\textbf{\boldmath $\fpr=10^{-1}$}} \\
\multicolumn{1}{c}{} & \multicolumn{1}{c}{} & \multicolumn{1}{c}{} & \multicolumn{1}{c}{} & \boldmath{($\delta=10^{-5}$)} & \multicolumn{1}{c}{} & \textbf{\boldmath $\overline{\Delta\tpr}$} & \textbf{\boldmath $p$} & \textbf{\boldmath $p_{\text{adjusted}}$}  & \textbf{\boldmath $\overline{\Delta\tpr}$} & \textbf{\boldmath $p$} & \textbf{\boldmath $p_{\text{adjusted}}$} & \textbf{\boldmath $\overline{\Delta\tpr}$} & \textbf{\boldmath $p$} & \textbf{\boldmath $p_{\text{adjusted}}$} \\
\midrule
\multirow{24}{*}{CIFAR10} & \multirow{6}{*}{ViT-B} & \multirow{6}{*}{Head} & \multirow{6}{*}{100} &\multirow{2}{*}{$\infty$} & ACC-LiRA & -71.79 & 0.993 & 1.00 & -67.26 & 0.980 & 1.00 & 114.11 & \textcolor{red}{0.0001} & 0.064 \\
 &  &  &  & & KL-LiRA & 72.88 & 0.0672 & 1.00 & -91.84 & 0.999 & 1.00 & -196.45 & 1.00 & 1.00 \\
\cmidrule{5-15}
 &  &  &  & \multirow{2}{*}{$8$} & ACC-LiRA & 3.14 & 0.070 & 1.00 & 8.71 & 0.112 & 1.00 & 14.77 & 0.083 & 1.00 \\
 &  &  &  &  & KL-LiRA & -2.31 & 0.930 & 1.00 & -10.28 & 0.874 & 1.00 & -11.32 & 0.639 & 1.00 \\
\cmidrule{5-15}
 &  &  &  & \multirow{2}{*}{$1$} & ACC-LiRA & 0.07 & 0.460 & 1.00 & 0.36 & 0.441 & 1.00 & -2.69 & 0.709 & 1.00 \\
 &  &  &  &  & KL-LiRA & 0.85 & 0.117 & 1.00 & -2.56 & 0.850 & 1.00 & 7.08 & 0.150 & 1.00 \\
\cmidrule{2-15}
 &  \multirow{6}{*}{ViT-B} & \multirow{6}{*}{Head} & \multirow{6}{*}{50} &\multirow{2}{*}{$\infty$}& ACC-LiRA & -1.51 & 0.542 & 1.00 & -18.30 & 0.780 & 1.00 & -32.01 & 0.801 & 1.00 \\
 &  &  &  & & KL-LiRA & 6.92 & 0.453 & 1.00 & 2.18 & 0.482 & 1.00 & 18.87 & 0.360 & 1.00 \\
\cmidrule{5-15}
 &  &  &  & \multirow{2}{*}{$8$} & ACC-LiRA & 1.82 & 0.198 & 1.00 & 3.36 & 0.347 & 1.00 & 16.67 & 0.225 & 1.00 \\
 &  &  &  &  & KL-LiRA & -2.95 & 0.878 & 1.00 & 2.71 & 0.365 & 1.00 & 21.63 & 0.141 & 1.00 \\
\cmidrule{5-15}
 &  &  &  & \multirow{2}{*}{$1$} & ACC-LiRA & 0.62 & 0.189 & 1.00 & -5.47 & 0.955 & 1.00 & -3.59 & 0.619 & 1.00 \\
 &  &  &  &  & KL-LiRA & 0.62 & 0.223 & 1.00 & 1.39 & 0.305 & 1.00 & -6.10 & 0.736 & 1.00 \\

\cmidrule{2-15}
 & \multirow{6}{*}{R-50} & \multirow{6}{*}{Head} & \multirow{6}{*}{50} &\multirow{2}{*}{$\infty$} & ACC-LiRA & 265.79 & \textcolor{red}{0.014} & 1.00 & 102.17 & 0.133 & 1.00 & -14.75 & 0.643 & 1.00 \\
 &  &  &  & & KL-LiRA & 81.36 & 0.080 & 1.00 & 4.18 & 0.435 & 1.00 & -0.47 & 0.506 & 1.00 \\
\cmidrule{5-15}
 &  &  &  & \multirow{2}{*}{$8$} & ACC-LiRA & -1.88 & 0.782 & 1.00 & -6.40 & 0.906 & 1.00 & -26.47 & 0.988 & 1.00 \\
 &  &  &  &  & KL-LiRA & -1.74 & 0.800 & 1.00 & -10.24 & 0.902 & 1.00 & -14.22 & 0.907 & 1.00 \\
\cmidrule{5-15}
 &  &  &  & \multirow{2}{*}{$1$} & ACC-LiRA & -1.50 & 0.907 & 1.00 & -3.52 & 0.842 & 1.00 & -4.65 & 0.778 & 1.00 \\
 &  &  &  &  & KL-LiRA & -0.50 & 0.761 & 1.00 & -2.61 & 0.946 & 1.00 & -4.18 & 0.705 & 1.00 \\

\cmidrule{2-15}
 &  \multirow{6}{*}{R-50} & \multirow{6}{*}{FiLM} & \multirow{6}{*}{$50$} & \multirow{2}{*}{$\infty$} & ACC-LiRA & 96.81 & 0.087 & 1.00 & 144.23 & \textcolor{red}{0.016} & 1.00 & 80.98 & 0.055 & 1.00 \\
 &  &  &  & & KL-LiRA & 5.58 & 0.469 & 1.00 & -3.97 & 0.530 & 1.00 & -36.75 & 0.754 & 1.00 \\
\cmidrule{5-15}
 &  &  &  & \multirow{2}{*}{$8$} & ACC-LiRA & -0.04 & 0.509 & 1.00 & -6.09 & 0.791 & 1.00 & -4.03 & 0.584 & 1.00 \\
 &  &  &  &  & KL-LiRA & -2.60 & 0.862 & 1.00 & -5.02 & 0.732 & 1.00 & -5.48 & 0.695 & 1.00 \\
\cmidrule{5-15}
 &  &  &  & \multirow{2}{*}{$1$} & ACC-LiRA & -0.65 & 0.744 & 1.00 & -4.90 & 0.868 & 1.00 & -1.88 & 0.585 & 1.00 \\
 &  &  &  &  & KL-LiRA & 1.05 & 0.147 & 1.00 & 1.43 & 0.385 & 1.00 & 14.50 & 0.120 & 1.00 \\
 
\cmidrule{1-15}
\multirow{6}{*}{CIFAR100} & \multirow{6}{*}{ViT-B} & \multirow{6}{*}{Head} & \multirow{6}{*}{100} & \multirow{2}{*}{$\infty$} & ACC-LiRA & -1.38 & 0.724 & 1.00 & 6.53 & 0.209 & 1.00 & 15.00 & 0.316 & 1.00 \\
 &  &  &  & & KL-LiRA & -22.41 & 0.810 & 1.00 & -22.05 & 0.814 & 1.00 & -18.82 & 0.780 & 1.00 \\
\cmidrule{5-15}
 &  &  &  & \multirow{2}{*}{$8$} & ACC-LiRA & -0.77 & 0.915 & 1.00 & -10.77 & 0.996 & 1.00 & -23.17 & 0.995 & 1.00 \\
 &  &  &  &  & KL-LiRA & 1.72 & 0.084 & 1.00 & 5.96 & 0.071 & 1.00 & 10.29 & 0.078 & 1.00 \\
\cmidrule{5-15}
 &  &  &  & \multirow{2}{*}{$1$} & ACC-LiRA & -0.04 & 0.562 & 1.00 & -0.04 & 0.524 & 1.00 & -1.37 & 0.733 & 1.00 \\
 &  &  &  &  & KL-LiRA & 0.09 & 0.306 & 1.00 & -1.06 & 0.922 & 1.00 & -0.96 & 0.680 & 1.00 \\

\bottomrule
\end{tabular}}
\end{center}
\end{table*}

\begin{table*}[htb]
\centering
\caption{Paired 1-sided permutation test (with $H_1: \overline{\tpr}_{\text{TD}} > \overline{\tpr}_{\text{ED}}$) table comparing $\mathbb{P}_{\text{TD}}(\tpr|\fpr)$ and $\mathbb{P}_{\text{ED}}(\tpr|\fpr)$. $\overline{\tpr}_{\text{TD}}$ denotes the average $\tpr$ when HPO for the target model is done using the training data (TD-HPO) and $\overline{\tpr}_{\text{ED}}$ denotes the average $\tpr$ when HPO is done using an external dat set (ED-HPO).   $\overline{\Delta\tpr}$ represents $(\overline{\tpr_{\text{TD}} - \tpr_{\text{ED}}}) \times 10^{-4}$.} 
\label{tab:permtest_table}
\begin{center}
\adjustbox{max width=\textwidth}{\begin{tabular}{lllrclccccccccc}
\toprule
\multicolumn{1}{c}{\multirow{2}{*}{\textbf{Dataset}}} & \multicolumn{1}{c}{\multirow{2}{*}{\textbf{Model}}} & \multicolumn{1}{c}{\multirow{2}{*}{\textbf{Config}}} & \multicolumn{1}{c}{\multirow{2}{*}{\textbf{\boldmath $S$}}} &  \multicolumn{1}{c}{\textbf{\boldmath $\varepsilon$}}  & \multicolumn{1}{c}{\multirow{2}{*}{\textbf{MIA}}} & \multicolumn{3}{c}{\textbf{\boldmath $\fpr=10^{-3}$}} & \multicolumn{3}{c}{\textbf{\boldmath $\fpr=10^{-2}$}} & \multicolumn{3}{c}{\textbf{\boldmath $\fpr=10^{-1}$}} \\
\multicolumn{1}{c}{} & \multicolumn{1}{c}{} & \multicolumn{1}{c}{} & \multicolumn{1}{c}{} & \boldmath{($\delta=10^{-5}$)} & \multicolumn{1}{c}{} & \textbf{\boldmath $\overline{\Delta\tpr}$} & \textbf{\boldmath $p$} & \textbf{\boldmath $p_{\text{adjusted}}$}  & \textbf{\boldmath $\overline{\Delta\tpr}$} & \textbf{\boldmath $p$} & \textbf{\boldmath $p_{\text{adjusted}}$} & \textbf{\boldmath $\overline{\Delta\tpr}$} & \textbf{\boldmath $p$} & \textbf{\boldmath $p_{\text{adjusted}}$} \\
\midrule
\multirow{24}{*}{CIFAR10} & \multirow{6}{*}{ViT-B} & \multirow{6}{*}{Head} & \multirow{6}{*}{100} &\multirow{2}{*}{$\infty$} & ACC-LiRA & -71.79 & 0.990  & 1.00 & -67.26 & 0.980 & 1.00 & 114.11 & \textcolor{red}{0.0008} & 0.476 \\
 &  &  &  & & KL-LiRA & 72.88 & 0.071329 & 1.00& -91.84 & 0.999 & 1.00 & -196.45 & 1.00 & 1.00 \\
\cmidrule{5-15}
 &  &  &  & \multirow{2}{*}{$8$} & ACC-LiRA & 3.14 & 0.070 & 1.00& 8.71 & 0.113 & 1.00& 14.77 & 0.081 & 1.00\\
 &  &  &  &  & KL-LiRA & -2.31 & 0.927 & 1.00 & -10.28 & 0.867 & 1.00 & -11.32 & 0.557 & 1.00 \\
\cmidrule{5-15}
 &  &  &  & \multirow{2}{*}{$1$} & ACC-LiRA & 0.07 & 0.462 & 1.00 & 0.36 & 0.436 & 1.00 & -2.69 & 0.704 & 1.00 \\
 &  &  &  &  & KL-LiRA & 0.85 & 0.122 & 1.00& -2.56 & 0.855 & 1.00 & 7.08 & 0.149 & 1.00\\
\cmidrule{2-15}

 &  \multirow{6}{*}{ViT-B} & \multirow{6}{*}{Head} & \multirow{6}{*}{50} & \multirow{2}{*}{$\infty$} & ACC-LiRA & -1.51 & 0.553 & 1.00 & -18.30 & 0.785 & 1.00 & -32.01 & 0.806 & 1.00 \\
 &  &  &  & & KL-LiRA & 6.92 & 0.460 & 1.00 & 2.18 & 0.483 & 1.00 & 18.87 & 0.355 & 1.00 \\
\cmidrule{5-15}
 &  &  &  & \multirow{2}{*}{$8$} & ACC-LiRA & 1.82 & 0.200 & 1.00& 3.36 & 0.348 & 1.00 & 16.67 & 0.223 & 1.00\\
 &  &  &  &  & KL-LiRA & -2.95 & 0.883 & 1.00 & 2.71 & 0.364 & 1.00 & 21.63 & 0.138 & 1.00\\
\cmidrule{5-15}
 &  &  &  & \multirow{2}{*}{$1$} & ACC-LiRA & 0.62 & 0.182 & 1.00& -5.47 & 0.952 & 1.00 & -3.59 & 0.624 & 1.00 \\
 &  &  & &  & KL-LiRA & 0.62 & 0.217 & 1.00& 1.39 & 0.311 & 1.00 & -6.10 & 0.740 & 1.00 \\

\cmidrule{2-15}
 & \multirow{6}{*}{R-50} & \multirow{6}{*}{Head} & \multirow{6}{*}{50} & \multirow{2}{*}{$\infty$} & ACC-LiRA & 265.79 & \textcolor{red}{0.020} & 1.00& 102.17 & 0.136 & 1.00& -14.75 & 0.648 & 1.00 \\
 &  &  &  & & KL-LiRA & 81.36 & 0.078 & 1.00& 4.18 & 0.435 & 1.00 & -0.47 & 0.512 & 1.00 \\
\cmidrule{5-15}
 &  &  &  & \multirow{2}{*}{$8$} & ACC-LiRA & -1.88 & 0.786 & 1.00 & -6.40 & 0.899 & 1.00 & -26.47 & 0.989 & 1.00 \\
 &  &  &  &  & KL-LiRA & -1.74 & 0.807 & 1.00 & -10.24 & 0.905 & 1.00 & -14.22 & 0.905 & 1.00 \\
\cmidrule{5-15}
 &  &  &  & \multirow{2}{*}{$1$} & ACC-LiRA & -1.50 & 0.909 & 1.00 & -3.52 & 0.850 & 1.00 & -4.65 & 0.773 & 1.00 \\
 &  &  &  &  & KL-LiRA & -0.50 & 0.761 & 1.00 & -2.61 & 0.951 & 1.00 & -4.18 & 0.691 & 1.00 \\

\cmidrule{2-15}
 &  \multirow{6}{*}{R-50}  & \multirow{6}{*}{FiLM} & \multirow{6}{*}{50} & \multirow{2}{*}{$\infty$} & ACC-LiRA & 96.81 & 0.086 & 1.00& 144.23 & \textcolor{red}{0.016} & 1.00& 80.98 & 0.055 & 1.00\\
 &  &  &  & & KL-LiRA & 5.58 & 0.468 & 1.00 & -3.97 & 0.541 & 1.00 & -36.75 & 0.749 & 1.00 \\
\cmidrule{5-15}
 &  &  &  & \multirow{2}{*}{$8$} & ACC-LiRA & -0.04 & 0.514 & 1.00 & -6.09 & 0.795 & 1.00 & -4.03 & 0.576 & 1.00 \\
 &  &  &  &  & KL-LiRA & -2.60 & 0.843 & 1.00 & -5.02 & 0.732 & 1.00 & -5.48 & 0.689 & 1.00 \\
\cmidrule{5-15}
 &  &  &  & \multirow{2}{*}{$1$} & ACC-LiRA & -0.65 & 0.756 & 1.00 & -4.90 & 0.868 & 1.00 & -1.88 & 0.586 & 1.00 \\
 &  &  &  &  & KL-LiRA & 1.05 & 0.143 & 1.00& 1.43 & 0.386 & 1.00 & 14.50 & 0.123 & 1.00\\
\cmidrule{1-15}
\multirow{6}{*}{CIFAR100} & \multirow{6}{*}{ViT-B} & \multirow{6}{*}{Head} & \multirow{6}{*}{100} &\multirow{2}{*}{$\infty$}& ACC-LiRA & -1.38 & 0.730 & 1.00 & 6.53 & 0.210 & 1.00& 15.00 & 0.336 & 1.00 \\
 &  &  &  & & KL-LiRA & -22.41 & 0.822 & 1.00 & -22.05 & 0.814 & 1.00 & -18.82 & 0.773 & 1.00 \\
\cmidrule{5-15}
 &  &  &  & \multirow{2}{*}{$8$} & ACC-LiRA & -0.77 & 0.917 & 1.00 & -10.77 & 0.994 & 1.00 & -23.17 & 0.995 & 1.00 \\
 &  &  &  &  & KL-LiRA & 1.72 & 0.087 & 1.00& 5.96 & 0.081 & 1.00& 10.29 & 0.080 & 1.00\\
\cmidrule{5-15}
 &  &  &  & \multirow{2}{*}{$1$} & ACC-LiRA & -0.04 & 0.558 & 1.00 & -0.04 & 0.524 & 1.00 & -1.37 & 0.728 & 1.00 \\
 &  &  &  &  & KL-LiRA & 0.09 & 0.303 & 1.00 & -1.06 & 0.921 & 1.00 & -0.96 & 0.668 & 1.00 \\
\bottomrule
\end{tabular}}
\end{center}
\end{table*}

\begin{figure*}
    \centering
    \includegraphics[width=\linewidth]{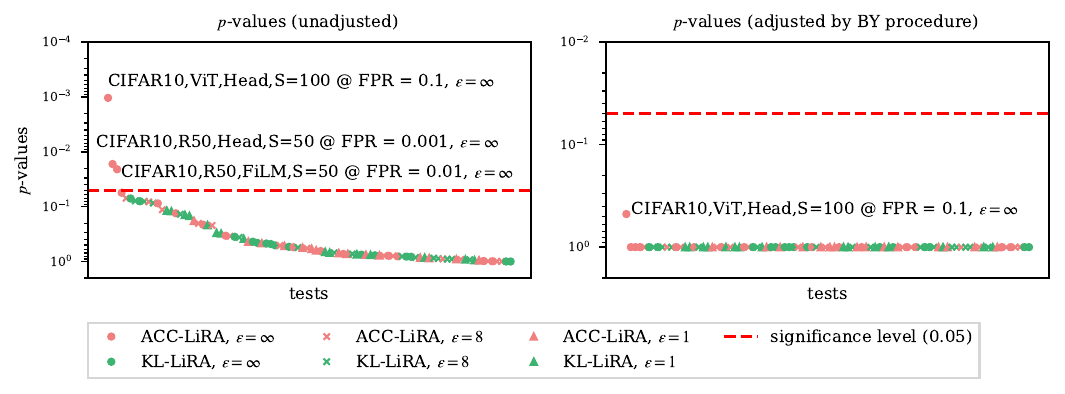}
    \caption{The plot above shows $p$-values (unadjusted and adjusted by Benjamini-Yekutieli (BY) procedure) on a negative logarithmic scale obtained for the permutation tests comparing the distribution of $\tpr$ at fixed $\fpr (0.001,0.01,0.1)$ for TD-HPO (where HPO is done using the training data), $\prob_{\text{TD}}(\tpr|\fpr)$, and ED-HPO (where HPO is done using external data), $\prob_{\text{ED}}(\tpr|\fpr)$, for all the experiments sorted in increasing order. The negative-logarithmic scale highlights the $p$-values that are significant. Additionally,
    we provide the details of the experiments for which the $p$-value is above the significance level or stands out among the adjusted $p$-values.}
    \label{fig:mahatten_plots_permtest_v2}.
\end{figure*}

\begin{figure*}
    \centering     
    \includegraphics[width=0.66\linewidth]{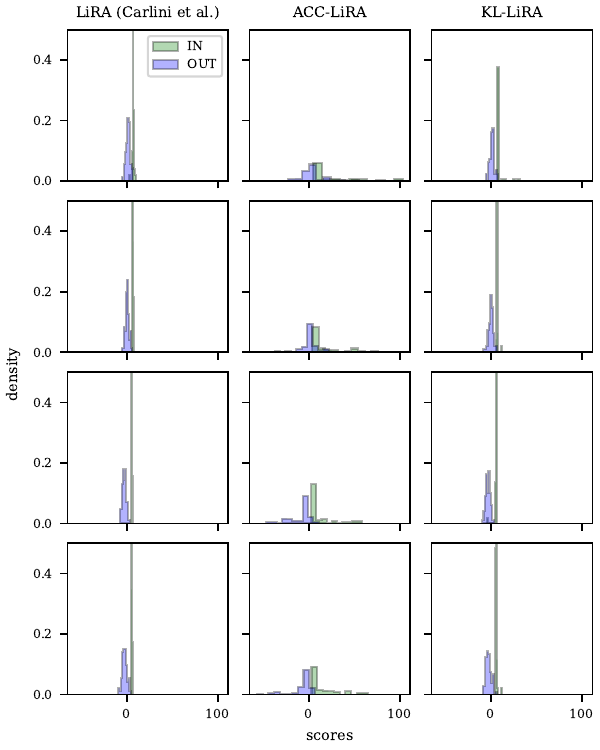}
    \caption{Samplewise IN and OUT score distributions for different attacks for samples drawn from CIFAR-100. The IN and OUT distributions for ACC-LiRA happen to show considerable overlapping compared to the corresponding score distributions for LiRA or KL-LiRA.}
    \label{fig:loss_dist_plot}
\end{figure*}

\begin{figure*}[htb]
    \centering
    \subfloat[CIFAR10, ViT + Head, $S=50$]{
    \includegraphics[width=0.9\textwidth]{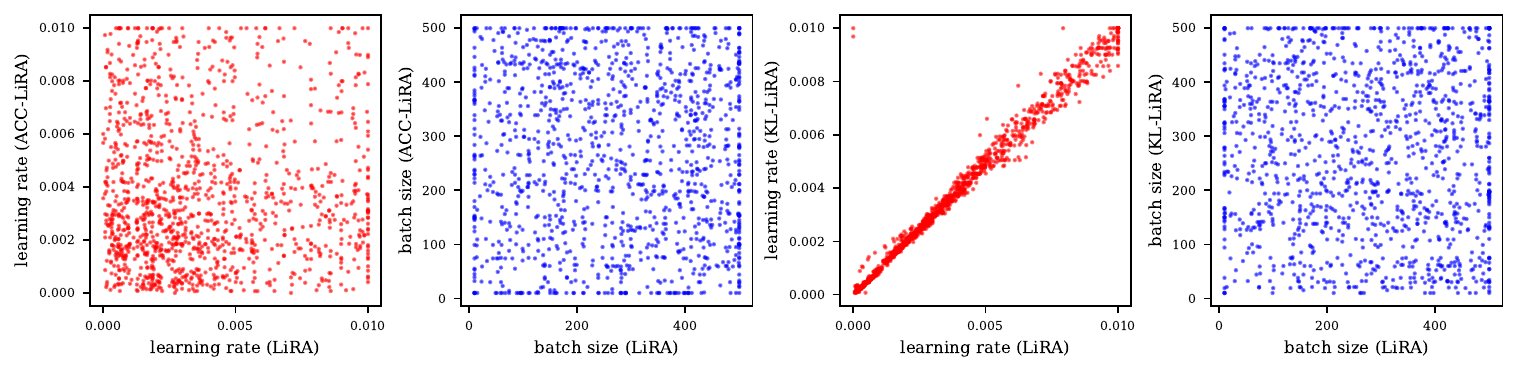}}
    
    \subfloat[CIFAR-100, ViT + Head, $S=100$]{
    \includegraphics[width=0.9\textwidth]{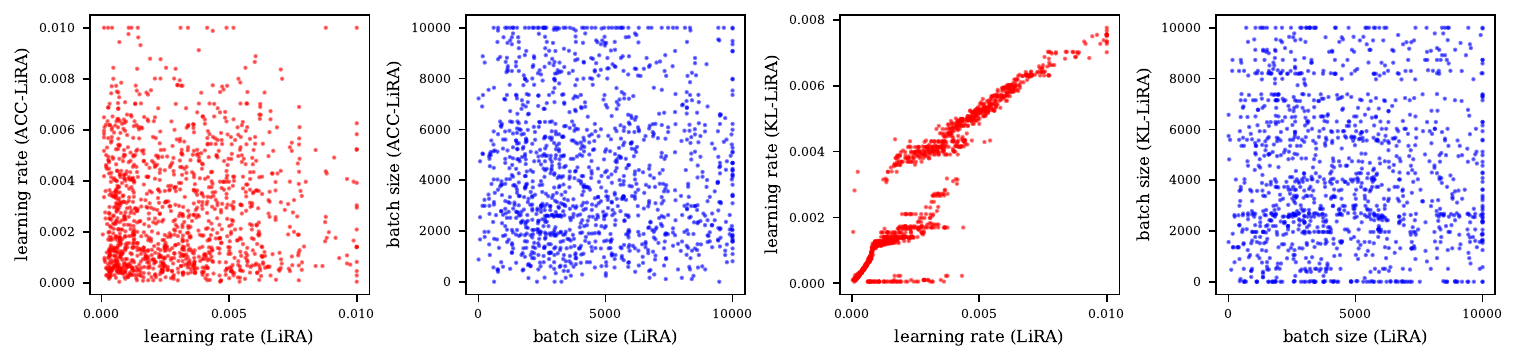}}
    
    \subfloat[CIFAR-10, R-50 + Head, $S=50$]{
    \includegraphics[width=0.9\textwidth]{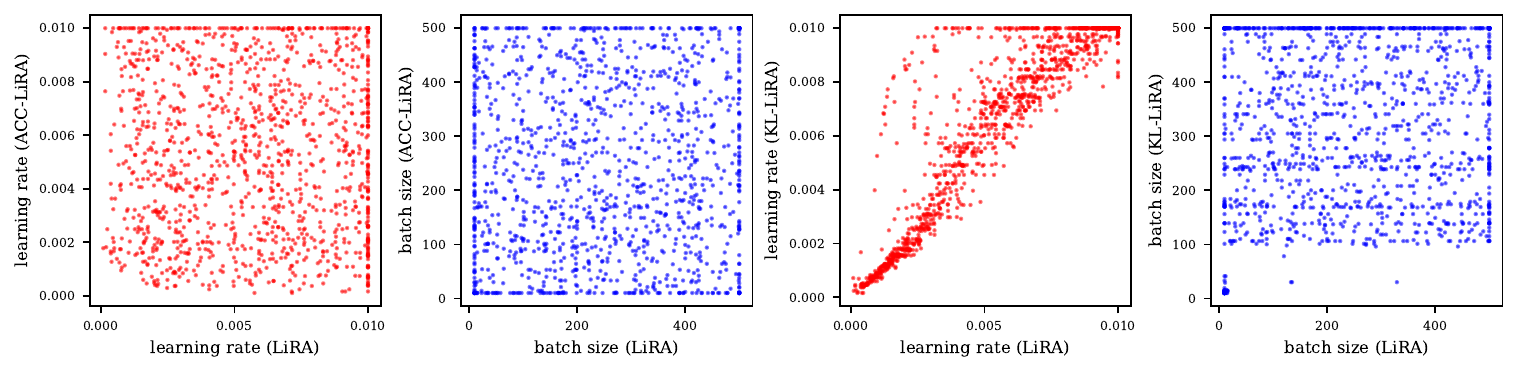}}
    
     \subfloat[CIFAR-10, R-50 + FiLM, $S=50$]{
    \includegraphics[width=0.9\textwidth]{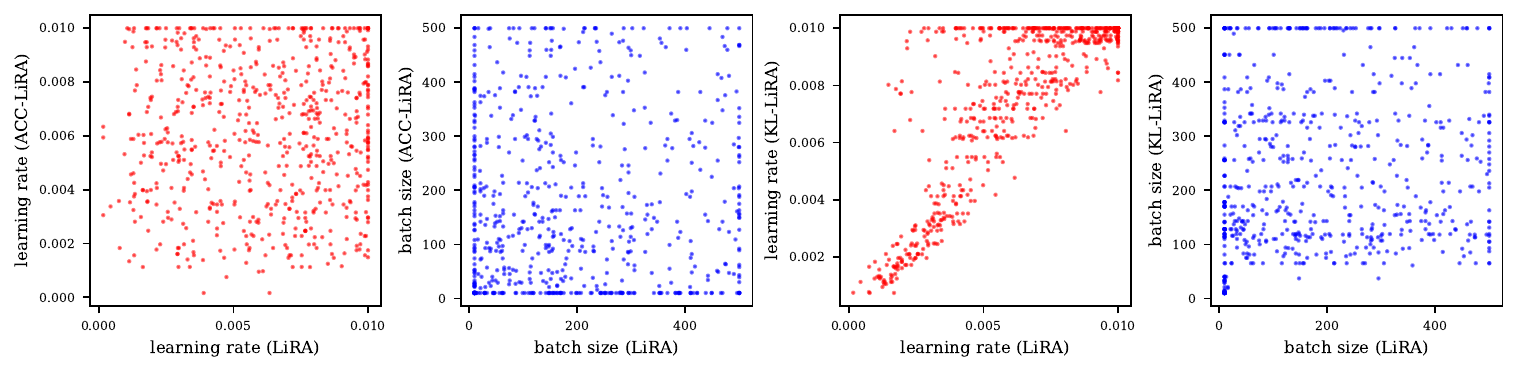}}
   \caption{ Scatter plots depicting relation between the hyperparameters used for LiRA (target model's hyperparameters) and the hyperparameters found using ACC-LiRA and KL-LiRA to train the shadow models. Learning rate found using KL-LiRA is close to the learning rate of the target model. For ACC-LiRA, neither the learning rate nor the batch size seems to have any meaningful relationship with the target model's hyperparameters as observed by the lack of a definite pattern in the corresponding scatter plots. Same is true for the relationship between batch size of the shadow models and the target models in KL-LiRA setting.}
    \label{fig:relation_hps_mias}
\end{figure*}

\end{document}